\documentclass[sigconf, nonacm]{acmart}

\usepackage{multirow}
\usepackage{hyperref}
\usepackage{amsmath}

\newcommand\vldbdoi{XX.XX/XXX.XX}
\newcommand\vldbpages{XXX-XXX}
\newcommand\vldbvolume{14}
\newcommand\vldbissue{1}
\newcommand\vldbyear{2020}
\newcommand\vldbauthors{\authors}
\newcommand\vldbtitle{\shorttitle} 
\newcommand\vldbavailabilityurl{URL_TO_YOUR_ARTIFACTS}
\newcommand\vldbpagestyle{plain} 

\begin{document}
\title{FECAM: Frequency Enhanced Channel Attention Mechanism for Time Series Forecasting}

%%
%% The "author" command and its associated commands are used to define the authors and their affiliations.
\author{Maowei Jiang$\dagger$} \thanks{$\dagger$These authors contributed equally to this work.}
\affiliation{%
  \institution{Shenyang Institute of Automation, Chinese Academy of Sciences}
  \streetaddress{P.O. Box 1212}
  \city{Shenyang}
  \state{China}
  \postcode{43017-6221}
}
\affiliation{%
  \institution{University of Chinese Academy of Sciences}
  \streetaddress{P.O. Box 1212}
  \city{Beijing}
  \state{China}
  \postcode{43017-6221}
}
\email{jiangmaowei@sia.cn}

\author{Pengyu Zeng$\dagger$}
\affiliation{%
  \institution{Shenyang Institute of Automation, Chinese Academy of Sciences}
  \streetaddress{P.O. Box 1212}
  \city{Shenyang}
  \state{China}
  \postcode{43017-6221}
}
\affiliation{%
  \institution{University of Chinese Academy of Sciences}
  \streetaddress{P.O. Box 1212}
  \city{Beijing}
  \state{China}
  \postcode{43017-6221}
}
\email{zengpengyu@sia.cn}

\author{Kai Wang$^{\ast}$} \thanks{*Corresponding author.}
\affiliation{%
  \institution{Shenyang Institute of Automation, Chinese Academy of Sciences}
  \streetaddress{P.O. Box 1212}
  \city{Shenyang}
  \state{China}
  \postcode{43017-6221}
}
\email{wangkai@sia.cn}

\author{Huan Liu}
\affiliation{%
  \institution{Shenyang Institute of Automation, Chinese Academy of Sciences}
  \streetaddress{P.O. Box 1212}
  \city{Shenyang}
  \state{China}
  \postcode{43017-6221}
}
\email{liuhuan@sia.cn}

\author{Wenbo Chen}
\affiliation{%
  \institution{Shenyang Institute of Automation, Chinese Academy of Sciences}
  \streetaddress{P.O. Box 1212}
  \city{Shenyang}
  \state{China}
  \postcode{43017-6221}
}
\affiliation{%
  \institution{University of Chinese Academy of Sciences}
  \streetaddress{P.O. Box 1212}
  \city{Beijing}
  \state{China}
  \postcode{43017-6221}
}
\email{chenwenbo@sia.cn}

\author{Haoran Liu}
\affiliation{%
  \institution{Shenyang Institute of Automation, Chinese Academy of Sciences}
  \streetaddress{P.O. Box 1212}
  \city{Shenyang}
  \state{China}
  \postcode{43017-6221}
}
\affiliation{%
  \institution{University of Chinese Academy of Sciences}
  \streetaddress{P.O. Box 1212}
  \city{Beijing}
  \state{China}
  \postcode{43017-6221}
}
\email{liuhaoran@sia.cn}

\begin{abstract}
Time series forecasting is a long-standing challenge due to the real-world information is in various scenario (e.g., energy, weather, traffic,economics, earthquake warning). However some mainstream forecasting model forecasting result is derailed dramatically from ground truth. we believe it’s the reason that models’ lacking ability of capturing frequency information which richly contains in real world datasets. At present, the mainstream frequency information extraction methods are Fourier transform(FT) based. However,use of FT is problematic due to Gibbs phenomenon. If the values on both sides of sequences differ significantly, oscillatory approximations are observed around both sides and high frequency noise will be introduced. Therefore We propose a novel frequency enhanced channel attention that adaptively modelling frequency interdependencies between channels based on Discrete Cosine Transform which would intrinsically avoid high frequency noise caused by problematic periodity during Fourier Transform, which is defined as Gibbs Phenomenon. We show that this network generalize extremely effectively across six real-world datasets and achieve state-of-the-art performance, we further demonstrate that frequency enhanced channel attention mechanism module can be flexibly applied to different networks. This module can improve the prediction ability of existing mainstream networks, which reduces \textbf{35.99\%} MSE on LSTM, \textbf{10.01\%} on Reformer, \textbf{8.71\%} on Informer, \textbf{8.29\%} on Autoformer, \textbf{8.06\%} on Transformer, etc., at a slight computational cost,with just a few line of code. Our codes and data are available at \href{https://github.com/Zero-coder/FECAM}{https://github.com/Zero-coder/FECAM}.
\end{abstract}

\maketitle

\pagestyle{\vldbpagestyle}
\begingroup\small\noindent\raggedright\textbf{PVLDB Reference Format:}\\
\vldbauthors. \vldbtitle. PVLDB, \vldbvolume(\vldbissue): \vldbpages, \vldbyear.\\
\href{https://doi.org/\vldbdoi}{doi:\vldbdoi}
\endgroup
\begingroup
\renewcommand\thefootnote{}\footnote{\noindent
This work is licensed under the Creative Commons BY-NC-ND 4.0 International License. Visit \url{https://creativecommons.org/licenses/by-nc-nd/4.0/} to view a copy of this license. For any use beyond those covered by this license, obtain permission by emailing \href{mailto:info@vldb.org}{info@vldb.org}. Copyright is held by the owner/author(s). Publication rights licensed to the VLDB Endowment. \\
\raggedright Proceedings of the VLDB Endowment, Vol. \vldbvolume, No. \vldbissue\ %
ISSN 2150-8097. \\
\href{https://doi.org/\vldbdoi}{doi:\vldbdoi} \\
}\addtocounter{footnote}{-1}\endgroup

\ifdefempty{\vldbavailabilityurl}{}{
\vspace{.3cm}
\begingroup\small\noindent\raggedright\textbf{PVLDB Artifact Availability:}\\
The source code, data, and/or other artifacts have been made available at \url{https://github.com/Zero-coder/FECAM}.
\endgroup
}

\section{Introduction}

Time series forecasting (TSF) enables decision-making with the estimated future evolution of metrics or events, thereby playing a crucial role in various scientific and engineering fields such as weather forecasting \cite{rasp2020weatherbench,grover2015deep,gneiting2005weather}, estimation of future illness cases \cite{ghassemi2015multivariate,reece2017forecasting,kandula2018evaluation}, energy consumption management \cite{zeng2022muformer,wang2019review,wei2019conventional,ruiz2018energy}, traffic flow \cite{ma2020hybrid,wong2020traffic,zhang2020spatio,zhang2018long},and financial investment \cite{barra2020deep,cao2019financial,wu2021dynamic,ding2020hierarchical}, to name a few.

With the growing data availability and computing power in recent years, it is shown that deep learning-based TSF methods can achieve much better prediction performance than traditional approaches \cite{liu2022SCINet}.

In recent year, Transformers \cite{NIPS2017_3f5ee243} have achieved progressive breakthrough on extensive areas \cite{chen2021decisiontransformer,Devlin2019BERTPO,dosovitskiy2021an,liu2021Swin}. Especially in time series forecasting, credited to their stacked structure and the capability of attention mechanisms, Transformers \cite{NIPS2017_3f5ee243,haoyietal-informer-2021,kitaev2020reformer} can naturally capture the temporal dependencies among time points, thereby fitting the series forecasting task perfectly.

\begin{figure*}[ht]
\centering
\includegraphics[width=0.9\linewidth]{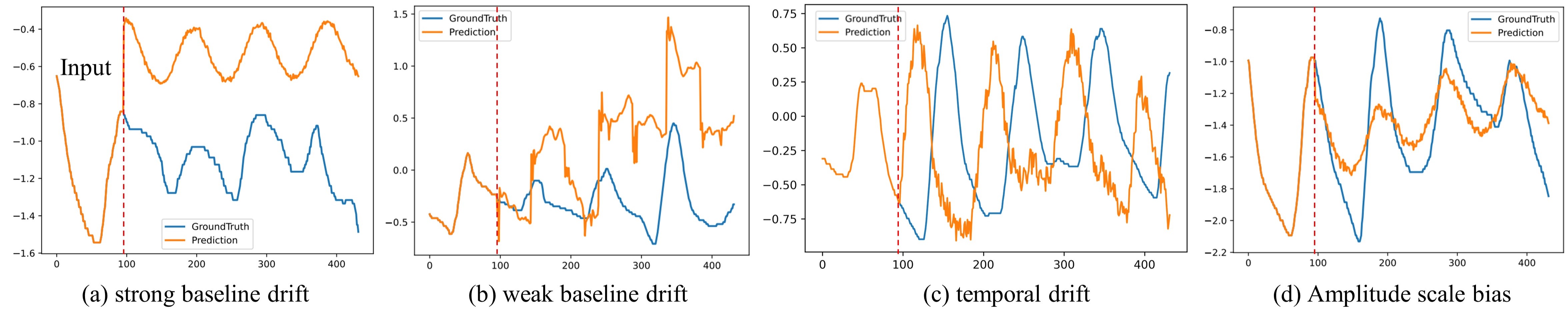}
\caption{The discrepancy between ground truth and forecasting output on real-world dataset ETTm2,(a) is from the vanilla LSTM,(b)is from vanilla Informer(c)is from vanilla transformer (d)is from vanilla autoformer.}
\label{Fig 1}
\end{figure*}

Despite the promising results of TSF methods, we found that the prediction of those methods,like transformers and LSTM is way derailed from the distribution of the ground truth of datasets, such as baseline drift in Fig.\ref{Fig 1}(a) and temporal drift in Fig.\ref{Fig 1}(c), we believe it’s the reason that models’ lacking ability of capturing frequency information which richly contains in real world datasets({Fig.\ref{33}), Therefore, the thing is that there still have room for improvement for these TSF mainstream methods to exploiting the natural property of time series data what we call frequency during modeling.

\begin{figure}[ht]
\centering
\includegraphics[width=1.0\linewidth]{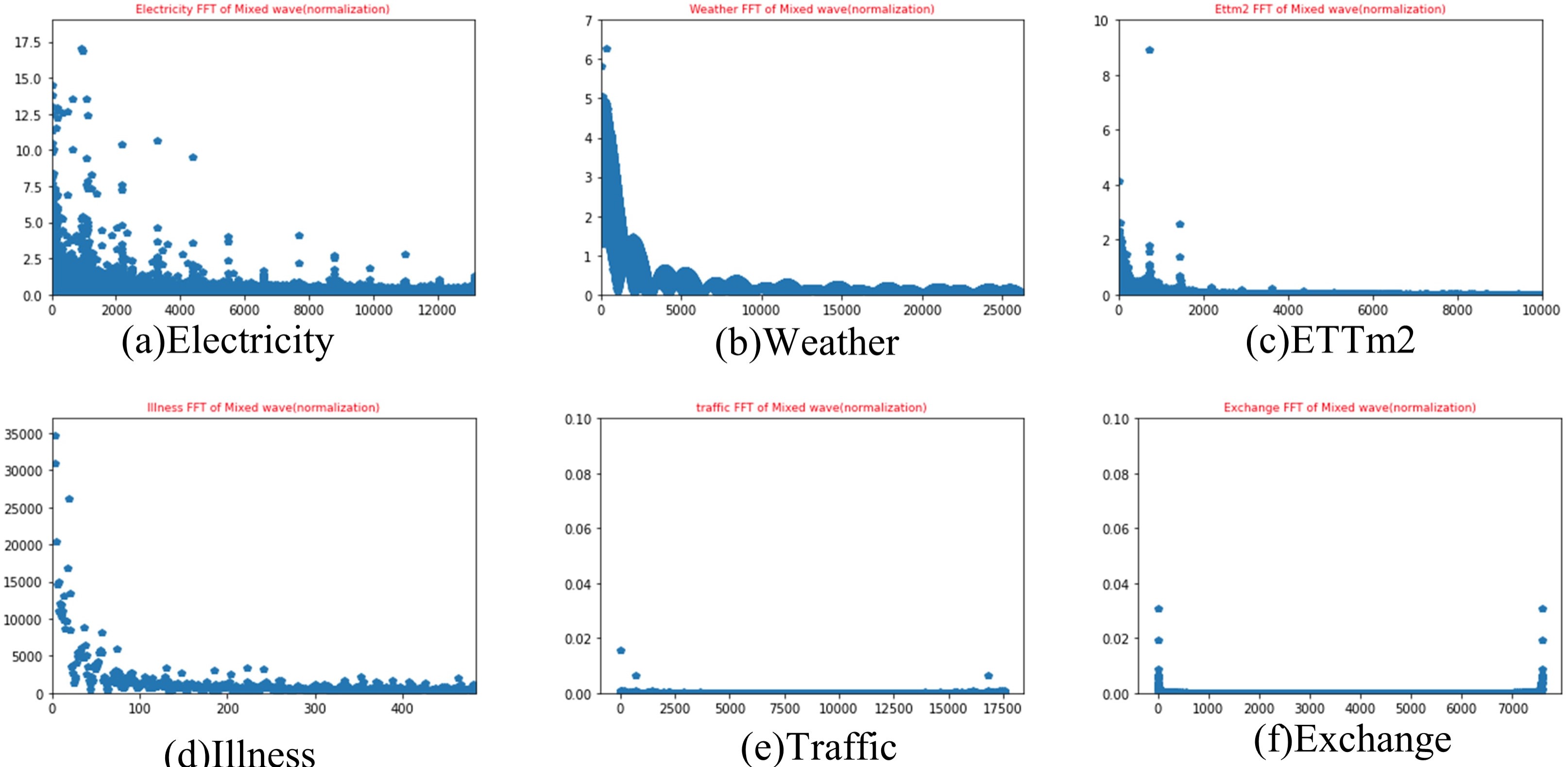}
\caption{six real world datasets visualization in Frequency domain, we can see most energy is contained in low frequency range.}
\label{33}
\end{figure}

Some efforts has been done for getting frequency representation and reconstructing temporal signal based on Fourier Transform and it's inverse transform. However, Fourier Transform (FT) would introduce high-frequency components for its problematic periodity, causing error value for boundary information which call Gibbs phenomenon and new round of computation consumption for inverse operation for avoiding complex operation in networks. Unlike FT/IFT based methods, Our method is based on Discrete Cosine Transform which would intrinsically eradicate Gibbs Phenomenon mentioned above and save unnecessary consumption of inverse transform, and for better exploiting utility of relationship between different time-series variate, we propose Frequency Enhanced Channel Attention Mechanism as a general framework, which empowers Transformer-based method and other mainstream models like LSTM, with better predictive ability for real-world time series.Consequently, effectively utilizing frequency information of time series enable us to perform forecasting with reasonable accuracy. Our method achieves state-of-the-art performance on six real-world benchmarks as a model. Furthermore,as a module FECAM can generalize to various Networks for further improvement, with just few line codes.

To this end, we propose a general feature extraction method for sequence modeling and forecasting, named frequency enhanced channel attention mechanism, which intrinsically eradicate Gibbs Phenomenon caused by Fourier Transform for the first time in time series forecasting.Our method achieves state-of-the-art performance on six real-world datasets, and can be generalized to other model architectures with just few line codes. The contributions of this paper are summarized as follows:

\begin{itemize}
\item We theoretically prove that our method can mitigate Gibbs phenomenon which would introduce high frequency noise during Fourier Transform,and we demonstrate that GAP is the lowest frequency component of DCT.
\item Based on above proof, We build the channel attention in frequency domain and propose our method with frequency enhanced channel mechanism for time-series forecasting. For generalization, we generalize frequency-enhanced channel attention into module that can be easily and flexibly adapted into other mainstream time series forecasting models to get better performance on six real-world datasets.
\item Extensive experiments on various TSF datasets show that FECAM as a general method consistently boosts four mainstream Transformers and non-transformer based methods like LSTM by a considerable margin and achieves state-of-the-art performance on six real-world datasets.
\end{itemize}

\section{Related Work and Preliminary}

\subsection{Deep Learning Models for Times series forecasting}

In recent years, deep learning models with meticulously designed architectures have achieved excellent progress in TSF tasks. RNN-based models \cite{Wen2017AMQ,2017Long, Maddix2018DeepFW, Rangapuram2018DeepSS, Flunkert2017DeepARPF} are proposed for application in an auto-regressive manner for sequence modeling, but the recurrent structure can suffer from problem of modeling long-term dependency. Shortly afterwards, Transformer \cite{NIPS2017_3f5ee243} emerges and shows great power in sequence modeling and gains great achievements in various downstream tasks. To solve the quadratic computation consumption on sequence length, subsequent works aim to decrease Self-Attention’s complexity. Particularly in long-term time series forecasting, Informer \cite{haoyietal-informer-2021} extends Self-Attention with KL-divergence criterion to select dominant queries. Reformer \cite{kitaev2020reformer} introduces local-sensitive hashing (LSH) mechanism to approximate attention by allocated similar queries. Not just improvement of reduction complexity, the following models further develop delicate building blocks for time series forecasting. Autoformer \cite{wu2021autoformer} coalesce the decomposition blocks into a canonical structure and designs Auto-Correlation to capture series-wise connections. Pyraformer \cite{liu2021pyraformer} designs pyramid attention module (PAM) to capture temporal dependencies with different hierarchies. Transformer-based models have taken the place of RNN-based models in almost all sequence modeling tasks, thanks to the effectiveness and efficiency of the self-attention mechanisms. Various Transformer-based TSF methods are proposed in the literature. These works typically focus on the challenging long-term time series forecasting problem, taking advantage of their remarkable long sequence modeling capabilities.

Although the transformers can capture long-range dependency in the time domain, it does not explicitly model the pattern occurrences in the frequency domain that plays an important role in tracking and predicting data points over various time cycles.

Different from previous works focusing on architectural design based on transformers, we analyze the series forecasting task from the natural view of frequency, which is the essential property of time series.It is also notable that as a general block, our proposed frequency-enhanced channel block can be easily applied to various models with a few operation. In the following subsection, we highlight our insights and motivate our work.

\subsection{Frequency Representation for time series forecasting}

Frequency is an indispensable information of time series, and real world datasets often contain rich frequency information as shown in Fig.\ref{33}, which allows better utilization of the capabilities of deep learning models. To utilize frequency information,Auto-former \cite{wu2021autoformer} use FFT in efficient computing of auto-correlation function, FNO \cite{Multiwavelet-based-Operator-Learning} is used as an inner block of networks to perform representation learning in low-frequency domain, DCTnet \cite{Xu_2020_CVPR} use Discrete Cosine Transform to compress information for keeping more original picture information in CV task. Most of these work was based on Fourier Transform which is helpful for extracting frequency features. However, most of the FT-based methods use Fourier Transform to get the frequency information and use Inverse Fourier Transform to reconstruct temporal information for avoiding complex-number training, which introduces new amount of computation, which is avoidable if using DCT for time-frequency transformation, what’s more the implicit periodicity of DFT gives rise to boundary discontinuity that result in significant high-frequency content which is known as Gibbson Phenomeneon. After quantization, Gibbs Phenomeneon causes the boundary points to take on erraneous values. SENET \cite{hu2018senet} only use GAP which is the lowest component of DFT and DCT for channel representation,meaning discarding other frequency-component information.

\subsection{Problem of Gibbs Phenomenon}

The Gibbs phenomenon \cite{SHIZGAL200341,foster1991gibbs,moskona1995gibbs} involves both the fact that Fourier sums overshoot at a jump discontinuity, and that this overshoot does not die out as more sinusoidal terms are added. And this would cause high frequencies noise which is supposed to be avoidable for time series forecasting. We demonstrate the phenomenon for square wave (In Fig.\ref{Gibs}) with the additive synthesis of a square wave with an increasing number of harmonics. The Gibbs phenomenon is visible especially when the number of harmonics is large. We give the mathematic description of Gibbs Phenomenon below.

\begin{figure}[ht]
\centering
\includegraphics[width=1.0\linewidth]{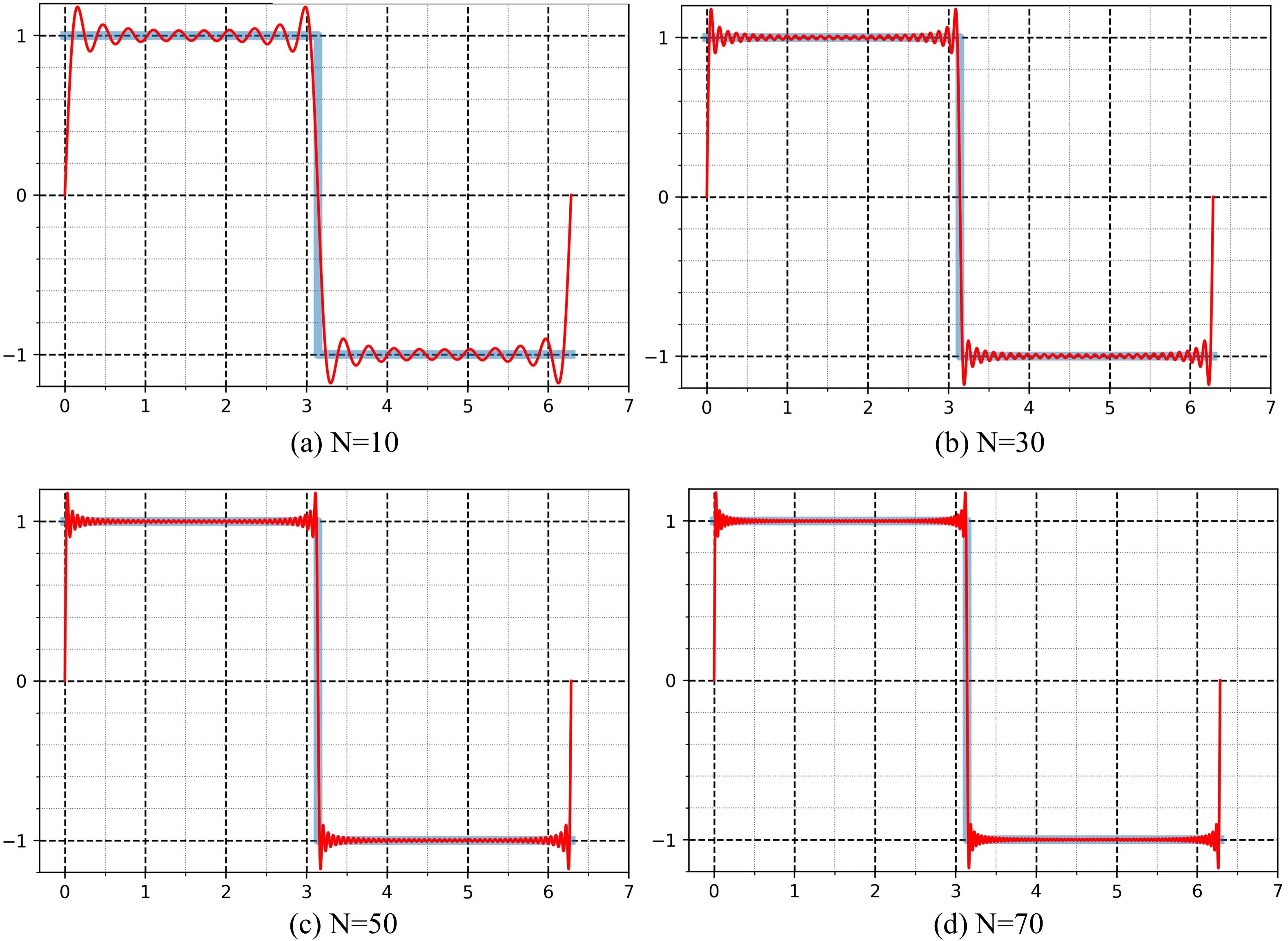}
\caption{Gibbs Phenomenon with increasing harmonics component.}
\label{Gibs}
\end{figure}

\begin{figure*}[ht]
\centering
\includegraphics[width=0.9\linewidth]{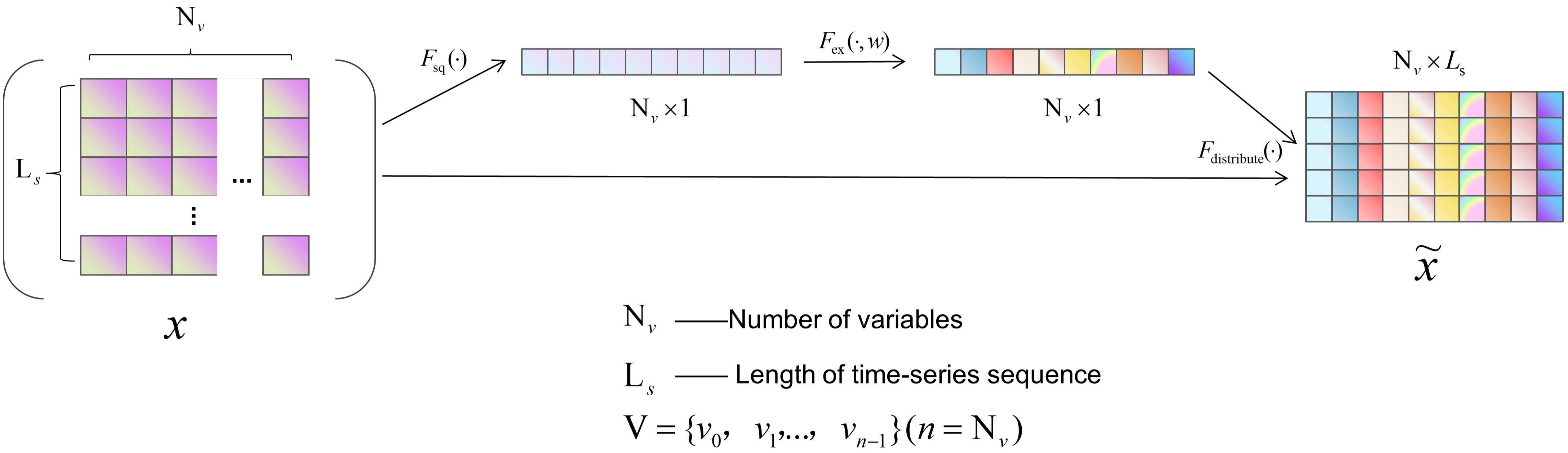}
\caption{SENET channel attention(Squeeze and excitation Network), Fsq(.) represent usg1d-global average pooling to extract global information a full connected layer for  from channel and redistribute weight for each channel.}
\label{11}
\end{figure*}

\textbf{Formal mathematical description of the phenomenon:} Let ${\displaystyle f:{\mathbb {R} }\to {\mathbb {R} }}$ be a piecewise continuously differentiable function which is periodic with some period ${\displaystyle L>0}$. Suppose that at some point ${\displaystyle x_{0}}$, the left limit ${\displaystyle f(x_{0}^{-})}$ and right limit ${\displaystyle f(x_{0}^{+})}$ of the function ${\displaystyle f}$ differ by a non-zero jump of ${\displaystyle a}$:

\begin{equation}
\begin{aligned}
{\displaystyle f(x_{0}^{+})-f(x_{0}^{-})=a\neq 0.}
\end{aligned}
\label{2.1}
\end{equation}

For each positive integer ${\displaystyle N}$, let ${\displaystyle S_{N}f(x)}$ be the ${\displaystyle N}$th partial Fourier series

\begin{equation}
\begin{aligned}
\displaystyle S_{N}f(x):&=\sum _{-N\leq n\leq N}{\widehat {f}}(n)e^{\frac {2i\pi nx}{L}}\\
&={\frac {1}{2}}a_{0}+\sum _{n=1}^{N}\left(a_{n}\cos \left({\frac {2\pi nx}{L}}\right)+b_{n}\sin \left({\frac {2\pi nx}{L}}\right)\right),
\end{aligned}
\label{2.2}
\end{equation}

where the Fourier coefficients ${\displaystyle {\widehat {f}}(n),a_{n},b_{n}}$ are given by the usual formulae

\begin{equation}
\begin{aligned}
\displaystyle {\widehat {f}}(n)&:={\frac {1}{L}}\int _{0}^{L}f(x)e^{-2i\pi nx/L}\,dx\\
\displaystyle a_{n}&:={\frac {2}{L}}\int _{0}^{L}f(x)\cos \left({\frac {2\pi nx}{L}}\right)\,dx\\
\displaystyle b_{n}&:={\frac {2}{L}}\int _{0}^{L}f(x)\sin \left({\frac {2\pi nx}{L}}\right)\,dx.
\end{aligned}
\label{2.3}
\end{equation}

Then we have:

\begin{equation}
\begin{aligned}
{\displaystyle \lim _{N\to \infty }S_{N}f\left(x_{0}+{\frac {L}{2N}}\right)=f(x_{0}^{+})+a\cdot (0.089489872236\dots )}
\end{aligned}
\label{2.4}
\end{equation}

and

\begin{equation}
\begin{aligned}
{\displaystyle \lim _{N\to \infty }S_{N}f\left(x_{0}-{\frac {L}{2N}}\right)=f(x_{0}^{-})-a\cdot (0.089489872236\dots )}
\end{aligned}
\label{2.5}
\end{equation}

but

\begin{equation}
\begin{aligned}
{\displaystyle \lim _{N\to \infty }S_{N}f(x_{0})={\frac {f(x_{0}^{-})+f(x_{0}^{+})}{2}}.}
\end{aligned}
\label{2.6}
\end{equation}

More generally, if ${\displaystyle x_{N}}$ is any sequence of real numbers which converges to ${\displaystyle x_{0}}$ as ${\displaystyle N\to \infty }$, and if the jump of ${\displaystyle a}$ is positive then

\begin{equation}
\begin{aligned}
{\displaystyle \limsup _{N\to \infty }S_{N}f(x_{N})\leq f(x_{0}^{+})+a\cdot (0.089489872236\dots )}
\end{aligned}
\label{2.7}
\end{equation}

and

\begin{equation}
\begin{aligned}
{\displaystyle \liminf _{N\to \infty }S_{N}f(x_{N})\geq f(x_{0}^{-})-a\cdot (0.089489872236\dots )}
\end{aligned}
\label{2.8}
\end{equation}

If instead the jump of ${\displaystyle a}$ is negative, one needs to interchange limit superior with limit inferior, and also interchange the ${\displaystyle \leq }$  and ${\displaystyle \geq }$ signs, in the above two inequalities.

\section{FECAM: Frequency Enhanced Channel Attention Mechanism}

Frequency is a natural auxiliary means to analyze time series. It is important and intuitive to introduce frequency information into time series models. However, most time series models tend to ignore the impact of frequency information on time series tasks, resulting in failure to learn the inherent characteristics of time series information. Most methods of extracting frequency information are based on FT and IFT, However,methods based on FT and IFT tends to introduce high-frequency noise due to problematic periodity which is known as Gibbs Phenomenon, Frequency Enhanced Frequency Channel Attention Mechanism can intrinsically avoid problem mentioned above and automatically acquire the importance of each channel through learning, it also suppresses features that are not useful for the current task.

We expect the learning of channel interdependencies features to be enhanced by explicitly modelling in frequency domain.

\subsection{Channel Attention and DCT}

We first elaborate on the definitions of discrete cosine transform and channel attention mechanism.

\subsubsection{Revisiting Channel attention}

The channel attention mechanism has been successfully introduced to CNNs. Squeeze-and-excitation (SE) block \cite{hu2018senet} models the interdependencies between the channels of feature maps with global information and recalibrate the feature maps to improve representation ability. It consists of squeeze and excitation two steps which are depicted in Fig .\ref{11}. For time-series signals $X\in R^{C\times L}$, $C$ is the number of channels, $L$ is the length of the temporal sequence, this type of tensor could be anywhere in the time-series model.

For temporal signals, the squeeze step applies GAP on temporal dimension to generate channel wise descriptor. Officially, a statistic $Z_{c}\in  R^{c}$ is generated by shrinking $x$ through its temporal dimension $L_S$, such that the $c$-th item of $z$ is calculated by: 

\begin{equation}
\begin{aligned}
Z_{c}=GAP(x_{c})=\frac{1}{L_{S}} \sum_{i=1}^{L_{S}} x_{c} (i) 
\end{aligned}
\label{1}
\end{equation}
Where $c, L_s$ represent the channel, and temporal dimension respectively. The scalar $Z_c$ is the $c$-th element of $Z$, Then the excitation step aims to modelling channel-wise dependencies by using two fully-connected layers $W_1$ and $W_2$ with a bottleneck architecture and non-linearity:
\begin{equation}
\begin{aligned}
att=\sigma (W_2\delta(W_1Z))
\end{aligned}
\label{2}
\end{equation}
where $att\in R^{C} $ is the learned attention vector which dot multiplies to the original feature map to re-scale each channel, $\sigma$ and $\delta$ refer to ReLU and sigmoid activation function respectively.

\begin{figure*}[ht]
\centering
\includegraphics[width=0.9\linewidth]{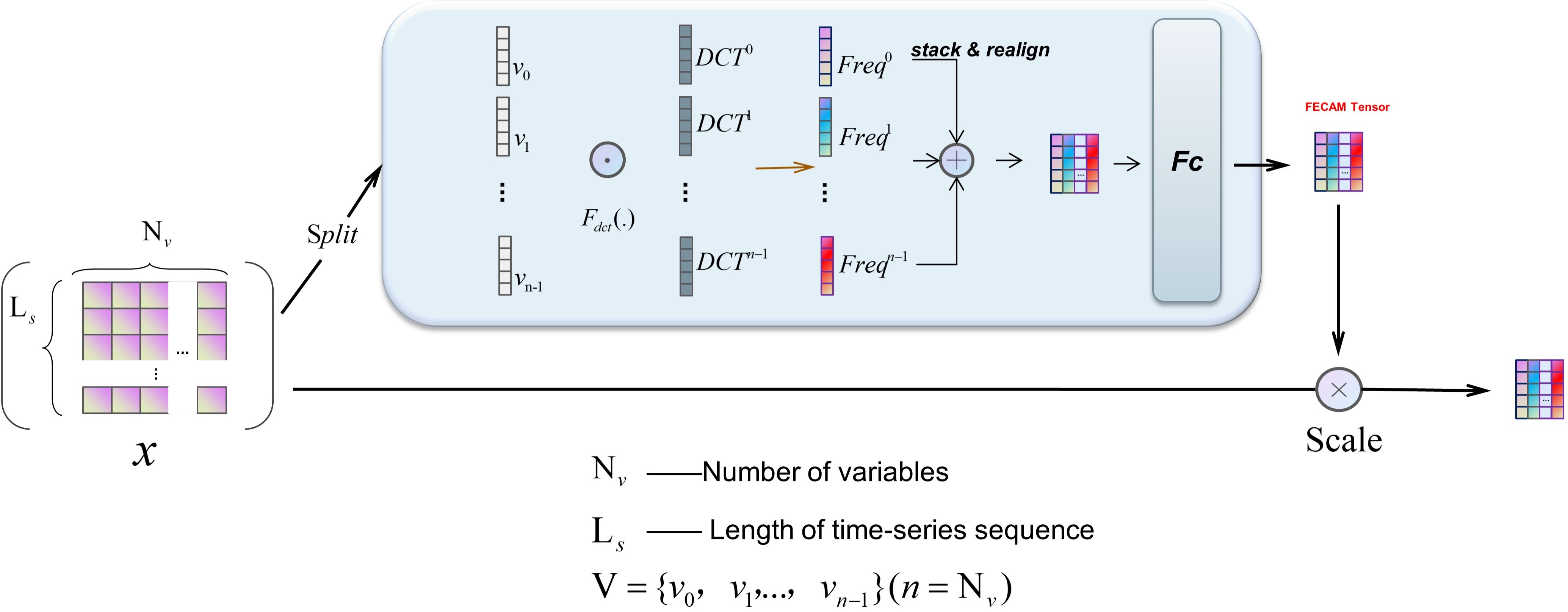}
\caption{Structure of Frequency Enhanced Channel Attention Mechanism. split every sequence of multivariate time series in each channel and Fdct(.) stands for Discrete Cosine Transform,stack\&realign each channel together.}
\label{22}
\end{figure*}

\subsubsection{Frequency representation for time series}
Sometimes, frequen-\\cy information contains more information that can be found, but it is difficult to mine in the time domain. For example, when a signal is disturbed by noise, its waveform will become messy, Or we can't distinguish it from noise in time domain. But it can be clearly distinguished from the frequency domain. Instead of well-known Fourier Transform,Our method introduce frequency information by Discrete Cosine Transform which can intrinsically avoid G-phenomenon and inverse transform operation.

\textbf{Discrete Cosine Transform (DCT)}

Typically, the basis function of one-dimensional (1D) DCT is:
\begin{equation}
\begin{aligned}
B_{l}^{i} =cos(\frac{\pi l}{L_s} (i+\frac{1}{2} ))
\end{aligned}
\label{3}
\end{equation}

Then the 1D DCT can be written as: 

\begin{equation}
\begin{aligned}
f_{1d}^{l} = {\textstyle \sum_{i=0}^{L_s-1}}x_{i}^{1d}  B_{l}^{i}
\end{aligned}
\label{4}
\end{equation}

s.t. $l\in \{0,1,\cdots ,L_s-1\}$,In which $f_{1d}^{l}\in R^L$ is the 1D DCT frequency spectrum, $x^{1d}\in R^L$ is the input, $L$ is the length of $x^{1d}$. Correspondingly, the inverse 1D DCT can be written as:
\begin{equation}
\begin{aligned}
x_{i}^{1d} = {\textstyle \sum_{i=0}^{L_s-1}}f_{l}^{1d}  B_{l}^{i}
\end{aligned}
\label{5}
\end{equation}

s.t. $i\in \{0,1,\cdots ,L_s-1\}$,In which $f_{1d}^{l}\in R^L$, Please note that in Eqs. 2 and 3, some constant normalization factors are removed for simplicity, which will not affect the results in this work.

\subsection{Frequency Enhanced Channel Attention Mechanism}

\begin{figure*}[ht]
\centering
\includegraphics[width=1.0\linewidth]{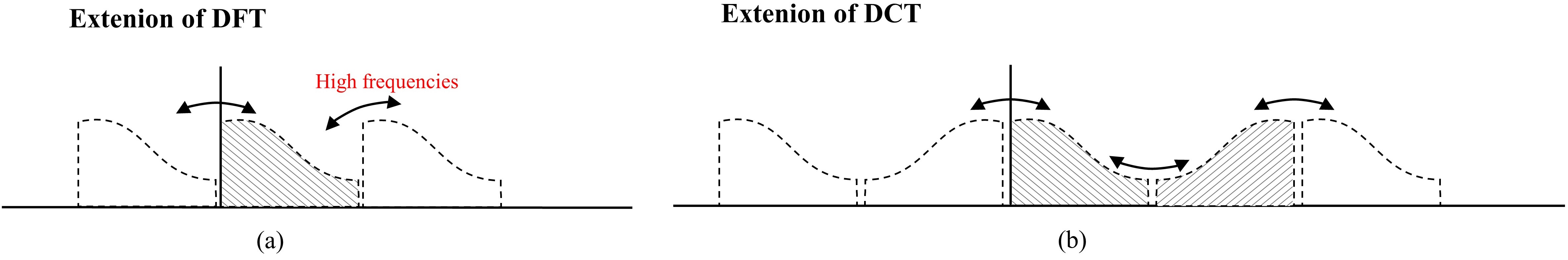}
\caption{Extension of Discrete Fourier Transform and Discrete Cosine Transform.}
\label{DCT}
\end{figure*}

In this section, we first theoretically discuss the problem of existing channel attention mechanisms. Based on the theoretical analysis, we then elaborate on the network design of the proposed method. 

Although GAP is a widely used operation in many attention mechanism as a standard squeezing method, we argue that simply use average-pooling on temporal dimension cause inadequate information extraction from time series which would even leads to information loss. Since GAP is the lowest frequency component of DCT and DFT, we mitigate this problem by introducing more frequency information. Rather than DFT, We use DCT to evade the Gibbs phenomenon mentioned many times before.
\\ \hspace*{\fill} \\
 \textbf{Theorem 1.} \textit{1d-GAP is a lowest component of 1D DCT, and its result is proportional to the lowest frequency component of 1d-DCT.}
\\ \hspace*{\fill} \\
\begin{equation}
\begin{aligned}
f_{0}^{1d}= {\textstyle \sum_{i=0}^{L_s-1}}x_{i}^{1d}cos\left ( \frac{0}{L_S}(i+\frac{1}{2} )\right)= {\textstyle \sum_{i=0}^{L_S-1}}x_{i}^{1d}=gap(x^{1d})L_S 
\end{aligned}
\label{6}
\end{equation}

In Eq.\ref{6}, $f_{0}^{1d}$ represents the lowest frequency component of 1D DCT, and it is proportional to GAP. In this way, Theorem 1 is proved. 

According to Theorem 1, without any surprise, we can sure that using GAP for feature extraction in channel attention means only the lowest frequency in obtained. All other frequency components are ignored, which supposed to be included in presenting channels.
\\ \hspace*{\fill} \\
 \textbf{Theorem 2.} \textit{Discrete Cosine Transform can intrinsically avoid Gibbs Phenomenon caused by periodic problem of Discrete Fourier Transform and Inverse Discrete Fourier Transform, and have a more efficient energy compaction than Fourier Transform.}
\\ \hspace*{\fill} \\
Discrete Cosine Transform is actually the DFT whose input signal is a real even function (proved in Derivation of DCT in Appendix A). Since Discrete Cosine Transform is using symmetric expansion for it’s periodic extension (In Fig.\ref{DCT}). Therefore, followed with eq.1, we have:

\begin{equation}
\begin{aligned}
f\left(x_0^{+}\right)-f\left(x_0^{-}\right)=a=0=f\left(x_0\right) \Leftrightarrow f\left(x_0^{+}\right)=f\left(x_0^{-}\right)
\end{aligned}
\label{3.1}
\end{equation}

Equation \ref{3.1} means that there’s no jump discontinuity which is necessary condition of Gibbs Phenomenon.

Then follow with eq.\ref{2.4} and eq.\ref{2.5},then we have:

\begin{equation}
\begin{aligned}
\lim _{N \rightarrow \infty} S_N f\left(x_0+\frac{L}{2 N}\right)=f\left(x_0^{+}\right)+0 \cdot(0.0894 \ldots)
\end{aligned}
\label{3.2}
\end{equation}

\begin{equation}
\begin{aligned}
\lim _{N \rightarrow \infty} S_N f\left(x_0-\frac{L}{2 N}\right)=f\left(x_0^{-}\right)-0 \cdot(0.0894 \ldots)
\end{aligned}
\label{3.3}
\end{equation}

Follow with eq.\ref{2.7} and eq.\ref{2.8},

\begin{equation}
\begin{aligned}
\Rightarrow \lim _{N \rightarrow \infty} \mathrm{S} u p S_N f\left(x_0\right) \leq f\left(x_0^{+}\right)
\end{aligned}
\label{3.4}
\end{equation}

\begin{equation}
\begin{aligned}
\Rightarrow \lim _{N \rightarrow \infty} \operatorname{Inf} S_N f\left(x_0\right) \geq f\left(x_0^{-}\right)
\end{aligned}
\label{3.5}
\end{equation}

\begin{equation}
\begin{aligned}
\lim _{N \rightarrow \infty} S_N f\left(x_0\right)=\frac{f\left(x_0^{+}\right)+f\left(x_0^{-}\right)}{2}=f\left(x_0\right)
\end{aligned}
\label{3.6}
\end{equation}

As we can see, the limit of the formula converges at this point, no oscillation observed, thus fundamentally eliminating the Gibbs effect. Because IFT and FT is consistent in mathematical nature, it is also true for the inverse DFT. It’s worthy to note that in the DFT case the periodic extension introduces discontinuities, which not happen for the DCT due to its property of symmetry extension, it is the elimination of this artificial discontinuity which contains a lot of high frequencies make the DCT is much more energy efficient than Discrete Fourier Transform. To this extent, Theorem 2 is proved. And we have done experiments to validate our Theorem 2 in section \ref{sec4.4}.

% \usepackage{multirow}
\begin{table*}[]
\centering
\renewcommand\arraystretch{1.1}
\caption{Statistics of datasets.}\smallskip
\aboverulesep= 0.8ex
\setlength{\tabcolsep}{11mm}{
\begin{tabular}{cccc}
\toprule[1.5pt]
\hline
Dataset     & Variable Number & Sampling Frequency & Total Observations \\ \hline
Exchange    & 8               & 1 Day              & 7,588              \\
ILI         & 7               & 1 Week             & 966                \\
ETTm2       & 7               & 15 Minutes         & 69,680             \\
Electricity & 321             & 1 Hour             & 26,304             \\
Traffic     & 862             & 1 Hour             & 17,544             \\
Weather     & 21              & 10 Minutes         & 52,695             \\ \hline \bottomrule[1.5pt]
\end{tabular}
\label{table Statistics}}
\end{table*}

% \usepackage{multirow}
\begin{table*}[]
\centering
\renewcommand\arraystretch{1.1}
\caption{Forecasting results comparison under different prediction lengths $O\in \{96,192,336,720\}$.The input sequence length is set to 36 for ILI and 96 for the others.}\smallskip
\aboverulesep= 0.8ex
\setlength{\tabcolsep}{2.1mm}{
\begin{tabular}{cccccccccccccccc}
\toprule[1.5pt]
\hline
\multicolumn{2}{c}{\multirow{2}{*}{Models}}                               & \multicolumn{2}{c}{\multirow{2}{*}{\textbf{Ours}}} & \multicolumn{2}{c}{\multirow{2}{*}{Autoformer}} & \multicolumn{2}{c}{\multirow{2}{*}{Pyraformer}} & \multicolumn{2}{c}{\multirow{2}{*}{Informer}} & \multicolumn{2}{c}{\multirow{2}{*}{LogTrans}} & \multicolumn{2}{c}{\multirow{2}{*}{Reformer}} & \multicolumn{2}{c}{\multirow{2}{*}{LSTNet}} \\
\multicolumn{2}{c}{}                                                      & \multicolumn{2}{c}{}                               & \multicolumn{2}{c}{}                            & \multicolumn{2}{c}{}                            & \multicolumn{2}{c}{}                          & \multicolumn{2}{c}{}                          & \multicolumn{2}{c}{}                          & \multicolumn{2}{c}{}                        \\ \cline{3-16} 
\multicolumn{2}{c}{\multirow{2}{*}{Metric}}                               & \multirow{2}{*}{MSE}     & \multirow{2}{*}{MAE}    & \multirow{2}{*}{MSE}   & \multirow{2}{*}{MAE}   & \multirow{2}{*}{MSE}   & \multirow{2}{*}{MAE}   & \multirow{2}{*}{MSE}  & \multirow{2}{*}{MAE}  & \multirow{2}{*}{MSE}  & \multirow{2}{*}{MAE}  & \multirow{2}{*}{MSE}  & \multirow{2}{*}{MAE}  & \multirow{2}{*}{MSE} & \multirow{2}{*}{MAE} \\
\multicolumn{2}{c}{}                                                      &                          &                         &                        &                        &                        &                        &                       &                       &                       &                       &                       &                       &                      &                      \\ \hline
\multicolumn{1}{c|}{\multirow{4}{*}{Exchange}} & \multicolumn{1}{c|}{96}  & \textbf{0.085}           & \textbf{0.208}          & 0.197                  & 0.323                  & 0.852                  & 0.780                  & 0.847                 & 0.752                 & 0.968                 & 0.812                 & 1.065                 & 0.829                 & 1.551                & 1.058                \\
\multicolumn{1}{c|}{}                          & \multicolumn{1}{c|}{192} & \textbf{0.210}           & \textbf{0.338}          & 0.300                  & 0.369                  & 0.993                  & 0.858                  & 1.204                 & 0.895                 & 1.04                  & 0.851                 & 1.610                 & 1.020                 & 1.477                & 1.028                \\
\multicolumn{1}{c|}{}                          & \multicolumn{1}{c|}{336} & \textbf{0.344}           & \textbf{0.445}          & 0.509                  & 0.524                  & 1.240                   & 0.958                  & 1.672                 & 1.036                 & 1.659                 & 1.081                 & 2.226                 & 1.192                 & 1.507                & 1.031                \\
\multicolumn{1}{c|}{}                          & \multicolumn{1}{c|}{720} & \textbf{0.921}           & \textbf{0.717}          & 1.447                  & 0.941                  & 1.711                  & 1.093                  & 2.478                 & 1.31                  & 1.941                 & 1.127                 & 1.802                 & 1.131                 & 2.285                & 1.243                \\ \hline
\multicolumn{1}{c|}{\multirow{4}{*}{ILI}}      & \multicolumn{1}{c|}{24}  & \textbf{2.101}           & \textbf{0.939}          & 3.483                  & 1.287                  & 5.800                  & 1.693                  & 5.764                 & 1.677                 & 4.475                 & 1.444                 & 4.400                 & 1.382                 & 6.026                & 1.770                \\
\multicolumn{1}{c|}{}                          & \multicolumn{1}{c|}{36}  & \textbf{2.330}           & \textbf{0.951}          & 3.103                  & 1.148                  & 6.043                  & 1.733                  & 4.755                 & 1.467                 & 4.799                 & 1.467                 & 4.783                 & 1.448                 & 5.340                & 1.668                \\
\multicolumn{1}{c|}{}                          & \multicolumn{1}{c|}{48}  & \textbf{2.557}           & \textbf{1.061}          & 2.669                  & 1.085                  & 6.213                  & 1.763                  & 4.763                 & 1.469                 & 4.800                   & 1.468                 & 4.832                 & 1.465                 & 6.080                & 1.787                \\
\multicolumn{1}{c|}{}                          & \multicolumn{1}{c|}{60}  & \textbf{2.531}           & \textbf{1.093}          & 2.770                  & 1.125                  & 6.531                  & 1.814                  & 5.264                 & 1.564                 & 5.278                 & 1.560                 & 4.882                 & 1.483                 & 5.548                & 1.720                \\ \hline
\multicolumn{1}{c|}{\multirow{4}{*}{ETTm2}}    & \multicolumn{1}{c|}{96}  & \textbf{0.188}           & \textbf{0.275}          & 0.255                  & 0.339                  & 0.409                  & 0.479                  & 0.365                 & 0.453                 & 0.768                 & 0.642                 & 0.658                 & 0.619                 & 3.142                & 1.365                \\
\multicolumn{1}{c|}{}                          & \multicolumn{1}{c|}{192} & \textbf{0.265}           & \textbf{0.336}          & 0.281                  & 0.340                  & 0.673                  & 0.641                  & 0.533                 & 0.563                 & 0.989                 & 0.757                 & 1.078                 & 0.827                 & 3.154                & 1.369                \\
\multicolumn{1}{c|}{}                          & \multicolumn{1}{c|}{336} & \textbf{0.318}           & \textbf{0.362}          & 0.339                  & 0.372                  & 1.210                  & 0.846                  & 1.363                 & 0.887                 & 1.334                 & 0.872                 & 1.549                 & 0.972                 & 3.160                & 1.369                \\
\multicolumn{1}{c|}{}                          & \multicolumn{1}{c|}{720} & \textbf{0.416}           & \textbf{0.417}          & 0.422                  & 0.419                  & 4.044                  & 1.526                  & 3.379                 & 1.388                 & 3.048                 & 1.328                 & 2.631                 & 1.242                 & 3.171                & 1.368                \\ \hline
\multicolumn{1}{c|}{\multirow{4}{*}{Electricity}}  & \multicolumn{1}{c|}{96}  & \textbf{0.178}           & \textbf{0.267}          & 0.201                  & 0.317                  & 0.498                  & 0.299                  & 0.274                 & 0.368                 & 0.258                 & 0.357                 & 0.312                 & 0.402                 & 0.680                & 0.645                \\
\multicolumn{1}{c|}{}                          & \multicolumn{1}{c|}{192} & \textbf{0.185}           & \textbf{0.273}          & 0.222                  & 0.334                  & 0.828                  & 0.312                  & 0.296                 & 0.386                 & 0.266                 & 0.368                 & 0.348                 & 0.433                 & 0.725                & 0.676                \\
\multicolumn{1}{c|}{}                          & \multicolumn{1}{c|}{336} & \textbf{0.199}           & \textbf{0.290}          & 0.231                  & 0.338                  & 1.476                  & 0.326                  & 0.300                 & 0.394                 & 0.280                 & 0.380                 & 0.350                 & 0.433                 & 0.828                & 0.727                \\
\multicolumn{1}{c|}{}                          & \multicolumn{1}{c|}{720} & \textbf{0.235}           & \textbf{0.323}          & 0.254                  & 0.361                  & 4.090                  & 0.372                  & 0.373                 & 0.439                 & 0.283                 & 0.376                 & 0.340                 & 0.420                 & 0.957                & 0.811                \\ \hline
\multicolumn{1}{c|}{\multirow{4}{*}{Traffic}}  & \multicolumn{1}{c|}{96}  & \textbf{0.493}           & \textbf{0.318}          & 0.613                  & 0.388                  & 0.684                  & 0.393                  & 0.719                 & 0.391                 & 0.684                 & 0.384                 & 0.732                 & 0.423                 & 1.107                & 0.685                \\
\multicolumn{1}{c|}{}                          & \multicolumn{1}{c|}{192} & \textbf{0.496}           & \textbf{0.319}          & 0.616                  & 0.382                  & 0.692                  & 0.394                  & 0.696                 & 0.379                 & 0.685                 & 0.390                 & 0.733                 & 0.420                 & 1.157                & 0.706                \\
\multicolumn{1}{c|}{}                          & \multicolumn{1}{c|}{336} & \textbf{0.511}           & \textbf{0.325}          & 0.622                  & 0.337                  & 0.699                  & 0.396                  & 0.777                 & 0.420                 & 0.733                 & 0.408                 & 0.742                 & 0.420                 & 1.216                & 0.730                \\
\multicolumn{1}{c|}{}                          & \multicolumn{1}{c|}{720} & \textbf{0.547}           & \textbf{0.343}          & 0.660                  & 0.408                  & 0.712                  & 0.404                  & 0.864                 & 0.472                 & 0.717                 & 0.396                 & 0.755                 & 0.423                 & 1.481                & 0.805                \\ \hline
\multicolumn{1}{c|}{\multirow{4}{*}{Weather}}  & \multicolumn{1}{c|}{96}  & \textbf{0.182}           & \textbf{0.242}          & 0.266                  & 0.336                  & 0.354                  & 0.392                  & 0.300                 & 0.384                 & 0.458                 & 0.490                 & 0.689                 & 0.596                 & 0.594                & 0.587                \\
\multicolumn{1}{c|}{}                          & \multicolumn{1}{c|}{192} & \textbf{0.223}           & \textbf{0.281}          & 0.307                  & 0.367                  & 0.673                  & 0.597                  & 0.598                 & 0.544                 & 0.658                 & 0.589                 & 0.752                 & 0.638                 & 0.560                & 0.565                \\
\multicolumn{1}{c|}{}                          & \multicolumn{1}{c|}{336} & \textbf{0.270}           & \textbf{0.320}          & 0.359                  & 0.395                  & 0.634                  & 0.592                  & 0.578                 & 0.523                 & 0.797                 & 0.652                 & 0.639                 & 0.596                 & 0.597                & 0.587                \\
\multicolumn{1}{c|}{}                          & \multicolumn{1}{c|}{720} & \textbf{0.338}           & \textbf{0.374}          & 0.419                  & 0.428                  & 0.942                  & 0.723                  & 1.059                 & 0.741                 & 0.869                 & 0.675                 & 1.130                 & 0.792                 & 0.618                & 0.599                \\ \hline \bottomrule[1.5pt]
\end{tabular}
\label{table 1}}
\end{table*}

% \usepackage{multirow}
\begin{table*}[]
\centering
\renewcommand\arraystretch{1.1}
\caption{Univariate results with different prediction lengths $O\in\{96,192,336,720\}$ on datasets ETTm2 and Exchange. The input sequence length is set to 96.}\smallskip
\aboverulesep= 0.8ex
\setlength{\tabcolsep}{1.5mm}{
\begin{tabular}{cccccccccccccccccc}
\toprule[1.5pt]
\hline
\multicolumn{2}{c}{\multirow{2}{*}{\textbf{Models}}}                      & \multicolumn{2}{c}{\multirow{2}{*}{\textbf{Ours}}} & \multicolumn{2}{c}{\multirow{2}{*}{N-HiTs}} & \multicolumn{2}{c}{\multirow{2}{*}{N-BEATS}} & \multicolumn{2}{c}{\multirow{2}{*}{Autoformer}} & \multicolumn{2}{c}{\multirow{2}{*}{Pyraformer}} & \multicolumn{2}{c}{\multirow{2}{*}{Informer}} & \multicolumn{2}{c}{\multirow{2}{*}{Reformer}} & \multicolumn{2}{c}{\multirow{2}{*}{ARIMA}} \\
\multicolumn{2}{c}{}                                                      & \multicolumn{2}{c}{}                               & \multicolumn{2}{c}{}                                 & \multicolumn{2}{c}{}                                  & \multicolumn{2}{c}{}                                     & \multicolumn{2}{c}{}                                     & \multicolumn{2}{c}{}                                   & \multicolumn{2}{c}{}                                   & \multicolumn{2}{c}{}                                \\ \cline{3-18} 
\multicolumn{2}{c}{\multirow{2}{*}{\textbf{Metric}}}                      & \multirow{2}{*}{MSE}     & \multirow{2}{*}{MAE}    & \multirow{2}{*}{MSE}      & \multirow{2}{*}{MAE}     & \multirow{2}{*}{MSE}      & \multirow{2}{*}{MAE}      & \multirow{2}{*}{MSE}        & \multirow{2}{*}{MAE}       & \multirow{2}{*}{MSE}        & \multirow{2}{*}{MAE}       & \multirow{2}{*}{MSE}       & \multirow{2}{*}{MAE}      & \multirow{2}{*}{MSE}       & \multirow{2}{*}{MAE}      & \multirow{2}{*}{MSE}     & \multirow{2}{*}{MAE}     \\
\multicolumn{2}{c}{}                                                      &                          &                         &                           &                          &                           &                           &                             &                            &                             &                            &                            &                           &                            &                           &                          &                          \\ \hline
\multicolumn{1}{c|}{\multirow{4}{*}{Exchange}} & \multicolumn{1}{c|}{96}  & \textbf{0.085}           & \textbf{0.208}          & 0.114                     & 0.248                    & 0.156                     & 0.299                     & 0.241                       & 0.387                      & 0.290                        & 0.439                      & 0.591                      & 0.615                     & 1.327                      & 0.944                     & 0.112                    & 0.245                    \\
\multicolumn{1}{c|}{}                          & \multicolumn{1}{c|}{192} & \textbf{0.206}           & \textbf{0.338}          & 0.250                     & 0.387                    & 0.669                     & 0.665                     & 0.273                       & 0.403                      & 0.594                       & 0.644                      & 1.183                      & 0.912                     & 1.258                      & 0.924                     & 0.304                    & 0.404                    \\
\multicolumn{1}{c|}{}                          & \multicolumn{1}{c|}{336} & \textbf{0.344}           & \textbf{0.445}          & 0.434                     & 0.516                    & 0.611                     & 0.605                     & 0.508                       & 0.539                      & 0.962                       & 0.824                      & 1.367                      & 0.984                     & 2.179                      & 1.296                     & 0.736                    & 0.598                    \\
\multicolumn{1}{c|}{}                          & \multicolumn{1}{c|}{720} & \textbf{0.759}           & \textbf{0.672}          & 1.061                     & 0.773                    & 1.111                     & 0.860                      & 0.991                       & 0.768                      & 1.285                       & 0.958                      & 1.872                      & 1.072                     & 1.280                       & 0.953                     & 1.871                    & 0.935                    \\ \hline
\multicolumn{1}{c|}{\multirow{4}{*}{ETTm2}}    & \multicolumn{1}{c|}{96}  & 0.066           & \textbf{0.188}          & 0.092                     & 0.232                    & 0.082                     & 0.219                     & \textbf{0.065}              & 0.189             & 0.074                       & 0.208                      & 0.088                      & 0.225                     & 0.131                      & 0.288                     & 0.211                    & 0.362                    \\
\multicolumn{1}{c|}{}                          & \multicolumn{1}{c|}{192} & \textbf{0.109}           & \textbf{0.245}          & 0.128                     & 0.276                    & 0,120                     & 0.268                     & 0.118                       & 0.256                      & 0.116                       & 0.252                      & 0.132                      & 0.283                     & 0.186                      & 0.354                     & 0.261                    & 0.406                    \\
\multicolumn{1}{c|}{}                          & \multicolumn{1}{c|}{336} & \textbf{0.144}           & \textbf{0.287}          & 0.165                     & 0.314                    & 0.226                     & 0.370                      & 0.154                       & 0.305                      & 0.143                       & 0.295                      & 0.180                      & 0.336                     & 0.220                       & 0.381                     & 0.317                    & 0.447                    \\
\multicolumn{1}{c|}{}                          & \multicolumn{1}{c|}{720} & \textbf{0.177}           & \textbf{0.326}          & 0.243                     & 0.397                    & 0.188                     & 0.338                     & 0.182                       & 0.335                      & 0.197                       & 0.338                      & 0.300                      & 0.435                     & 0.267                      & 0.430                     & 0.366                    & 0.487                    \\ \hline \bottomrule[1.5pt]
\end{tabular}
\label{table 2}}
\end{table*}

According to Theorem 2, we can found that in the DFT case the extension introduces discontinuities and this does not happen for DCT, due to the symmetry of its periodic extension, then our method eliminate this artificial discontinuity which contains a lot of high frequencies.

For capturing more time series information from feature map, we try to introduce DCT for getting more frequency components instead of only the GAP for lowest frequency \cite{hu2018senet}. Since DCT weight are constant, it can be pre-calculated only once and saved in advance, what’s more, results are real number, which means no training time for inverse transform and number of network parameters. Therefore,we propose frequency enhanced channel attention mechanism (FECAM) which can not only be used as a model for forecasting with just adding a projection layer but also can be seamlessly added to the existing time series forecasting models for improving their prediction performance. The overall structure of FECAM is shown in Fig.\ref{22}.

First, FECAM splits the input feature maps along the channel dimension into $n$ sub-groups as $[v_0,v_1,\cdots ,v_{n-1}]$, in which $V^i=R^{1\times L}$, $i\in \{0,1,\cdots,n-1 \},n=N_v$, Subsequently,for sub-group will be processed by a corresponding DCT frequency component ranging from low frequency to high frequency, Every single channel will processed by the same frequency component, In this way we have:

\begin{equation}
\begin{aligned}
Freq^i=DCT_j(V^i)= {\textstyle \sum_{j=0}^{j=L_S-1}(V_{:,l}^{i})B_{l}^{j}} 
\end{aligned}
\label{7}
\end{equation}

s.t. $i\in \{0,1,\cdots,N_V-1\}$, $j\in \{0,1,\cdots,L_S-1\}$, in which $j$ are the frequency component 1D indices corresponding to $V^i$,and $Freq^i\in R^L$ is the $L$ dimensional vector after the discrete cosine transformation. The whole  frequency channel vector can be obtained by stack operation.

\begin{equation}
\begin{aligned}
Freq=DCT(V)=stack([Freq^0,Freq^1,\cdots,Freq^{n-1}])
\end{aligned}
\label{8}
\end{equation}

In which $Freq\in R^{C\times L}$ is the attention vector for $V\in R^{C\times L}$. Once we obtain $Freq$, the attention weight can be learned through neural structure as SE-block. The whole frequency enhanced channel attention mechanism framework can be written as:

\begin{equation}
\begin{aligned}
F_c-att=\sigma \left ( W_2\delta(W_1DCT(V)) \right ) 
\end{aligned}
\label{9}
\end{equation}

By doing so, each channel features interact with every frequency components to acquire important temporal information comprehensively from frequency domain,which would encourages networks to enhance the diversity of extracted features. In the subsequent experiment section 4.3, we visualize the frequency channel attention tensor Fig.\ref{44}, demonstrating that FECAM learned the importance of different channels in the frequency domain and the importance of different frequency component pairs in each channel.

\section{Experiments}
We conduct extensive experiments to evaluate the performance of frequency enhanced channel mechanism network on six real-world time series forecasting benchmarks and further validate the generality of the proposed method on various mainstream Transformer variants and non-transformer based models.As a module embedding to other Networks,we have also done experiment of parameters increment and performance promotion and the visualization of frequency channel attention tensor to prove proposed method’s effectiveness and efficiency.

\textbf{Datasets:} Here are the descriptions of the datasets: 

\textbf{Electricity \footnote[1]{The Electricity dataset was acquired at https://archive.ics.uci.edu/ml/datasets/Electric-\\ityLoadDiagrams20112014}:} records the hourly electricity consumption of 321 clients from 2012 to 2014. 

\textbf{ETT \footnote[2]{The ETT dataset was acquired at https://github.com/zhouhaoyi/ETDataset}:} contains the time series of oil temperature and power load collected by electricity transformers from July 2016 to July 2018. ETTm1 /ETTm2 are recorded every 15 minutes, and ETTh1/ETTh2 are recorded every hour.

\textbf{Exchange \footnote[3]{The Exchange dataset was acquired at https://github.com/thuml/Autoformer}:} collects the panel data of daily exchange rates from 8 countries from 1990 to 2016.

\textbf{ILI \footnote[4]{The ILI dataset was acquired at https://gis.cdc.gov/grasp/fluview/fluportaldashboar-\\d.html}:} collects the ratio of influenza-like illness patients versus the total patients in one week,which is reported weekly by Centers for Disease Control and Prevention of the United States from2002 and 2021. 

\textbf{Traffic \footnote[5]{The Traffic dataset was acquired at http://pems.dot.ca.gov/}:} contains hourly road occupancy rates measured by 862 sensors on San Francisco Bay area freeways from January 2015 to December 2016. 

\textbf{Weather \footnote[6]{The Weather dataset was acquired at https://www.bgc-jena.mpg.de/wetter/}:} includes meteorological time series with 21 weather indicators collected every 10 minutes from the Weather Station of the Max Planck Biogeochemistry Institute in 2020. 

Table \ref{table Statistics} summarizes overall statistics of the datasets. We follow the standard protocol that divides each dataset into the training, validation, and testing subsets according to the chronological order. The split ratio is 3:1:1 for the ETT dataset and 7:2:2 for others.

\textbf{Baselines:} We evaluate the single full-connected layer equipped by the Frequency Enhanced Channel Attention mechanism in both multivariate and uni-variate settings to demonstrate its effectiveness. For multivariate forecasting, we include six state-of-the-art deep forecasting models: Autoformer \cite{wu2021autoformer}, Pyraformer \cite{liu2021pyraformer}, Informer \cite{haoyietal-informer-2021}, LogTrans \cite{2019Enhancing}, Reformer \cite{kitaev2020reformer} and LSTNet \cite{2018Modeling}. For univariateforecasting, we include seven competitive baselines:N-HiTS \cite{challu2022n},N-BEATS \cite{oreshkin2019n},Autoformer \cite{wu2021autoformer}, Pyraformer \cite{liu2021pyraformer}, Informer \cite{haoyietal-informer-2021}, Reformer \cite{kitaev2020reformer} and ARIMA \cite{Anderson1976TimeSeries2E}. In addition, we adopt the proposed framework on both the canonical and efficient variants of Transformers and classical RNNs:Transformer \cite{NIPS2017_3f5ee243},Informer \cite{haoyietal-informer-2021},Reformer \cite{kitaev2020reformer} and Autoformer \cite{wu2021autoformer} and LSTM \cite{greff2016lstm} to validate the generality of our framework.

\textbf{Implementation details:} All the experiments are implemented in PyTorch \cite{Paszke2019PyTorchAI} and conducted for three runs on a single NVIDIA GeForce RTX 3090 24GB GPU. Each model is trained by ADAM \cite{DBLP:journals/corr/KingmaB14} using L2 loss with the initial learning rate of 10e-4 and batch size of 32. Each Transformer-based model contains two encoder layers and one decoder layer. We report the test MSE/MAE under different prediction lengths as the performance metric. A lower MSE/MAE indicates better performance of time series forecasting.

\subsection{Main Results}

\textbf{Forecasting results:} As for multivariate forecasting results, Our proposed method with a projection layer for forecasting achieves state-of-the-art performance in all benchmarks and prediction leng-\\ths (Table \ref{table 1}). Notably, Frequency Enhanced Channel Attention Mechanism outperforms other deep models impressively characterized by much less model parameters. Compared with Autoformer, the proposed FECAM yields an \textbf{overall 21.52\%} relative MSE reduction and relative \textbf{10.78\%} MAE reduction. With the prediction length of  24 and 48,FECAM achieve an \textbf{39.67\%} MSE reduction(3.483→2.101) and \textbf{24.9\%}(3.103→2.330) respectively on ILI, with the the prediction length of {96,192,336,720}, FECAM achieve \textbf{36.40\%} relative MSE on Exchange compared to previous state-of-the-art results, which indicates that the potential of deep model is still constrained on ability of modelling in frequency domain. We also list the univariate results of two typical datasets with different frequency distribution (as shown in Fig.\ref{33}). FECAM still realizes remarkable forecasting performance.

% \usepackage{multirow}
\begin{table*}[]
\centering
\renewcommand\arraystretch{1.4}
\caption{Parameters increment and performance promotion of FECAM}\smallskip
\aboverulesep= 0.8ex
\setlength{\tabcolsep}{5.5mm}{
\begin{tabular}{cccccc}
\toprule[1.5pt]
\hline
\multicolumn{1}{l}{Models}    & LSTM    & Reformer & Informer & Autoformer & Transformer \\ \hline
Vanilla                       & 13.2K   & 5.79MB   & 11.33MB  & 10.54MB    & 10.54MB     \\ \hline
Vanilla+Ours                  & 13.5K   & 5.85MB   & 11.39MB  & 10.69MB    & 10.60MB     \\ \hline
Parameters increment          & \textbf{2.27\%}  & \textbf{1.03\%}   & \textbf{0.53\%}   & \textbf{0.57\%}     & \textbf{0.57\%}      \\ \hline
Performance promotion         & \textbf{35.99\%} & \textbf{10.01\%}  & \textbf{8.71\%}   & \textbf{8.29\%}     & \textbf{8.06\%}      \\ \hline \bottomrule[1.5pt]
\end{tabular}
\label{table 4}}
\end{table*}

% \usepackage{multirow}
\begin{table*}[]
\centering
\renewcommand\arraystretch{1.1}
\caption{Performance promotion by applying our proposed method  method to Transformers and RNNs.We report the averaged MSE/MAE of all prediction length (stated in Table \ref{table 1}) and the relative MSE reduction ratios(Promotion) by our method. Complete results can be found in Appendix B}\smallskip
\aboverulesep= 0.8ex
\setlength{\tabcolsep}{3.1mm}{
\begin{tabular}{ccccccccccccc}
\toprule[1.5pt]
\hline
Dataset        & \multicolumn{2}{c}{Exchange}         & \multicolumn{2}{c}{ILI}              & \multicolumn{2}{c}{ETTm2}            & \multicolumn{2}{c}{Electricity}      & \multicolumn{2}{c}{Traffic}          & \multicolumn{2}{c}{Weather}          \\
Model          & MSE               & MAE              & MSE               & MAE              & MSE               & MAE              & MSE               & MAE              & MSE               & MAE              & MSE               & MAE              \\ \hline
LSTM           & 2.104             & 1.221            & 6.537             & 1.828            & 2.394             & 1.177            & 0.559             & 0.549            & 1.010             & 0.541            & 0.443             & 0.453            \\
\textbf{+Ours} & 1.294             & 0.946            & 4.305             & 1.442            & 1.338             & 0.896            & 0.381             & 0.437            & 0.755             & 0.430            & 0.277             & 0.333            \\ \hline
promotion      & \multicolumn{2}{c}{\textbf{38.49\%}} & \multicolumn{2}{c}{\textbf{34.14\%}} & \multicolumn{2}{c}{\textbf{44.11\%}} & \multicolumn{2}{c}{\textbf{31.84\%}} & \multicolumn{2}{c}{\textbf{25.24\%}} & \multicolumn{2}{c}{\textbf{37.47\%}} \\ \hline
Transformer    & 1.556             & 0.969            & 4.774             & 0.445            & 1.344             & 0.814            & 0.272             & 0.367            & 0.667             & 0.363            & 0.681             & 0.576            \\
\textbf{+Ours} & 1.271             & 0.874            & 4.471             & 1.394            & 1.254             & 0.806            & 0.256             & 0.364            & 0.662             & 0.359            & 0.615             & 0.537            \\ \hline
promotion      & \multicolumn{2}{c}{\textbf{18.31\%}} & \multicolumn{2}{c}{\textbf{6.77\%}}  & \multicolumn{2}{c}{\textbf{6.69\%}}  & \multicolumn{2}{c}{\textbf{5.88\%}}  & \multicolumn{2}{c}{\textbf{0.75\%}}  & \multicolumn{2}{c}{\textbf{9.69\%}}  \\ \hline
Informer       & 1.550             & 0.998            & 5.136             & 1.544            & 1.410             & 0.822            & 0.31              & 0.396            & 0.764             & 0.415            & 0.633             & 0.548            \\
\textbf{+Ours} & 1.433             & 0.949            & 4.676             & 1.453            & 1.249             & 0.794            & 0.288             & 0.38             & 0.736             & 0.399            & 0.576             & 0.511            \\ \hline
promotion      & \multicolumn{2}{c}{\textbf{7.54\%}}  & \multicolumn{2}{c}{\textbf{8.95\%}}  & \multicolumn{2}{c}{\textbf{11.41\%}} & \multicolumn{2}{c}{\textbf{7.09\%}}  & \multicolumn{2}{c}{\textbf{3.66\%}}  & \multicolumn{2}{c}{\textbf{10.42\%}} \\ \hline
Autoformer     & 0.613             & 0.539            & 3.006             & 1.161            & 0.324             & 0.367            & 0.227             & 0.337            & 0.627             & 0.378            & 0.337             & 0.381            \\
\textbf{+Ours} & 0.504             & 0.499            & 2.738             & 1.108            & 0.315             & 0.359            & 0.217             & 0.326            & 0.616             & 0.367            & 0.318             & 0.368            \\ \hline
promotion      & \multicolumn{2}{c}{\textbf{17.78\%}} & \multicolumn{2}{c}{\textbf{8.91\%}}  & \multicolumn{2}{c}{\textbf{2.77\%}}  & \multicolumn{2}{c}{\textbf{4.40\%}}  & \multicolumn{2}{c}{\textbf{1.75\%}}  & \multicolumn{2}{c}{\textbf{5.63\%}}  \\ \hline
Reformer       & 1.620             & 1.023            & 4.724             & 1.445            & 1.479             & 0.915            & 0.337             & 0.422            & 0.740             & 0.421            & 0.802             & 0.655            \\
\textbf{+Ours} & 1.275             & 0.907            & 4.398             & 1.378            & 1.443             & 0.897            & 0.318             & 0.397            & 0.711             & 0.394            & 0.585             & 0.551            \\ \hline
promotion      & \multicolumn{2}{c}{\textbf{21.29\%}} & \multicolumn{2}{c}{\textbf{6.90\%}}  & \multicolumn{2}{c}{\textbf{2.43\%}}  & \multicolumn{2}{c}{\textbf{5.63\%}}  & \multicolumn{2}{c}{\textbf{3.91\%}}  & \multicolumn{2}{c}{\textbf{27.05\%}} \\ \hline \bottomrule[1.5pt]
\end{tabular}
\label{table 5}}
\end{table*}

\begin{figure}[ht]
\centering
\includegraphics[width=1.0\linewidth]{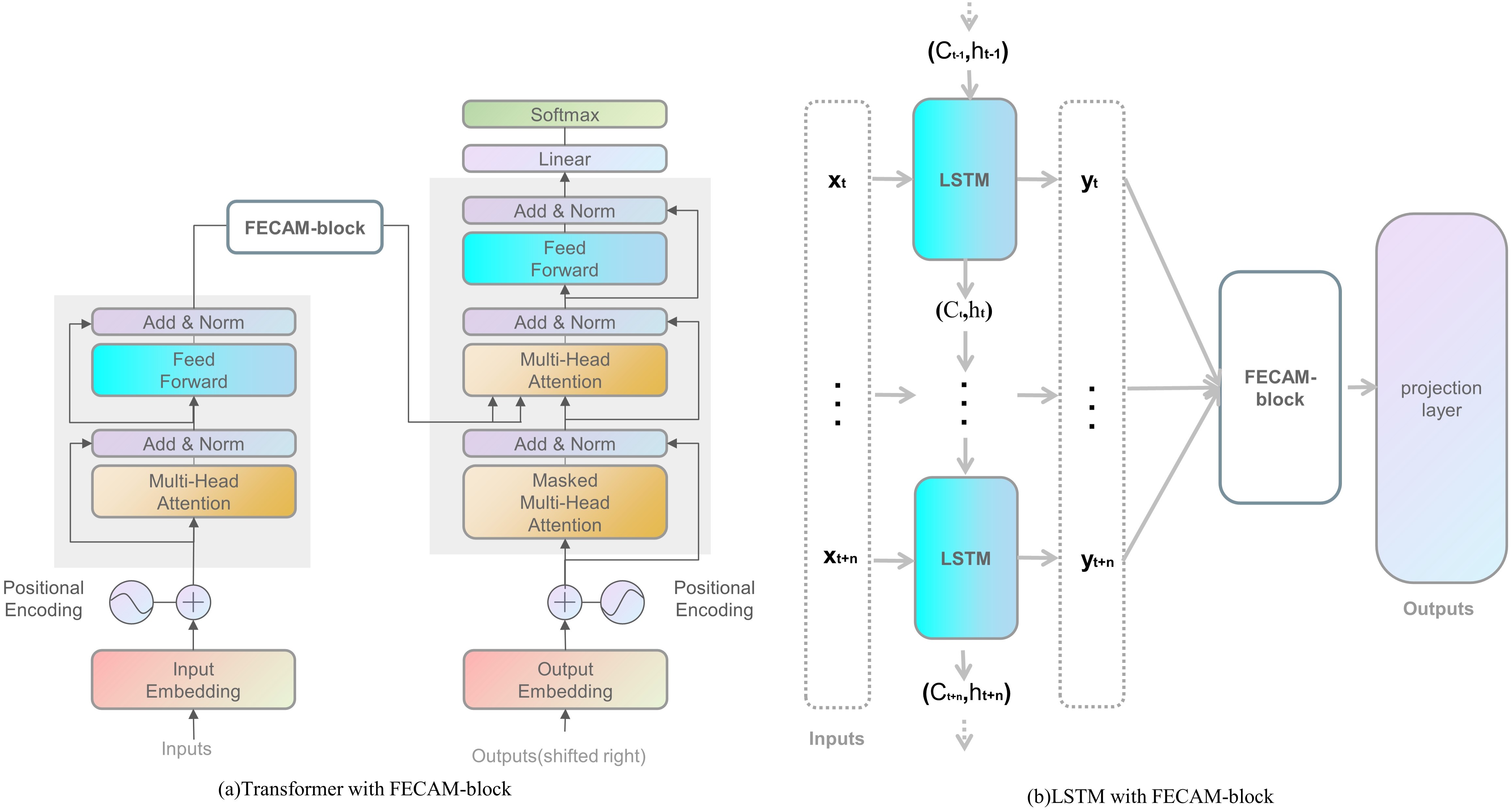}
\caption{FECAM as a module embedded into other Networks, figure (a) represent module is put behind the encoder of  Transformers and figure (b) represent module is put between LSTM output layer and projection layer.}
\label{LT}
\end{figure}

\textbf{Module generality:} We apply our proposed method to four mainstream Transformers (as shown in Fig.\ref{LT}(a)) and a mainstream recurrent neural network LSTM (as shown in Fig.\ref{LT}(b)) and report the performance promotion of each model (Table \ref{table 5}). Our method consistently improves the forecasting ability of different models. Overall, it achieves averaged 35.99\% promotion on LSTM, 10.01\% on Reformer, 8.71\% on Informer,8.29\% on Autoformer and 8.06\% on Transformer, making each of them surpass previous state-of-the-art. Compared to vanilla  models, only a few parameters  are increased by applying our method (See Table \ref{table 4}), and thereby their computational complexities can be preserved. It validates that Frequency Enhanced Channel Attention Mechanism is an effective and lightweight tool that can be widely applied to Transformer-based models and RNNs with a few line code, and enhances their ability of modelling in frequency domain to achieve state-of-the-art performance.

By analyzing the results of Table \ref{table 5}, we can obviously find that the module gains of the FECAM are large in datasets Exchange, ETTm2, and weather, but small in the dataset traffic. By observing the frequency spectrum of each dataset (as shown in the Fig.\ref{33}), we can safely say that datasets like Exchange, ETTm2, and Weather have a lot of energy information at low frequencies, while there is little energy information in the frequency spectrum of the Traffic dataset, ant this might be the reason why the gains of FECAM module is not so profitable on the traffic dataset.

\subsection{Model Analysis}

\begin{figure}[ht]
\centering
\includegraphics[width=1.0\linewidth]{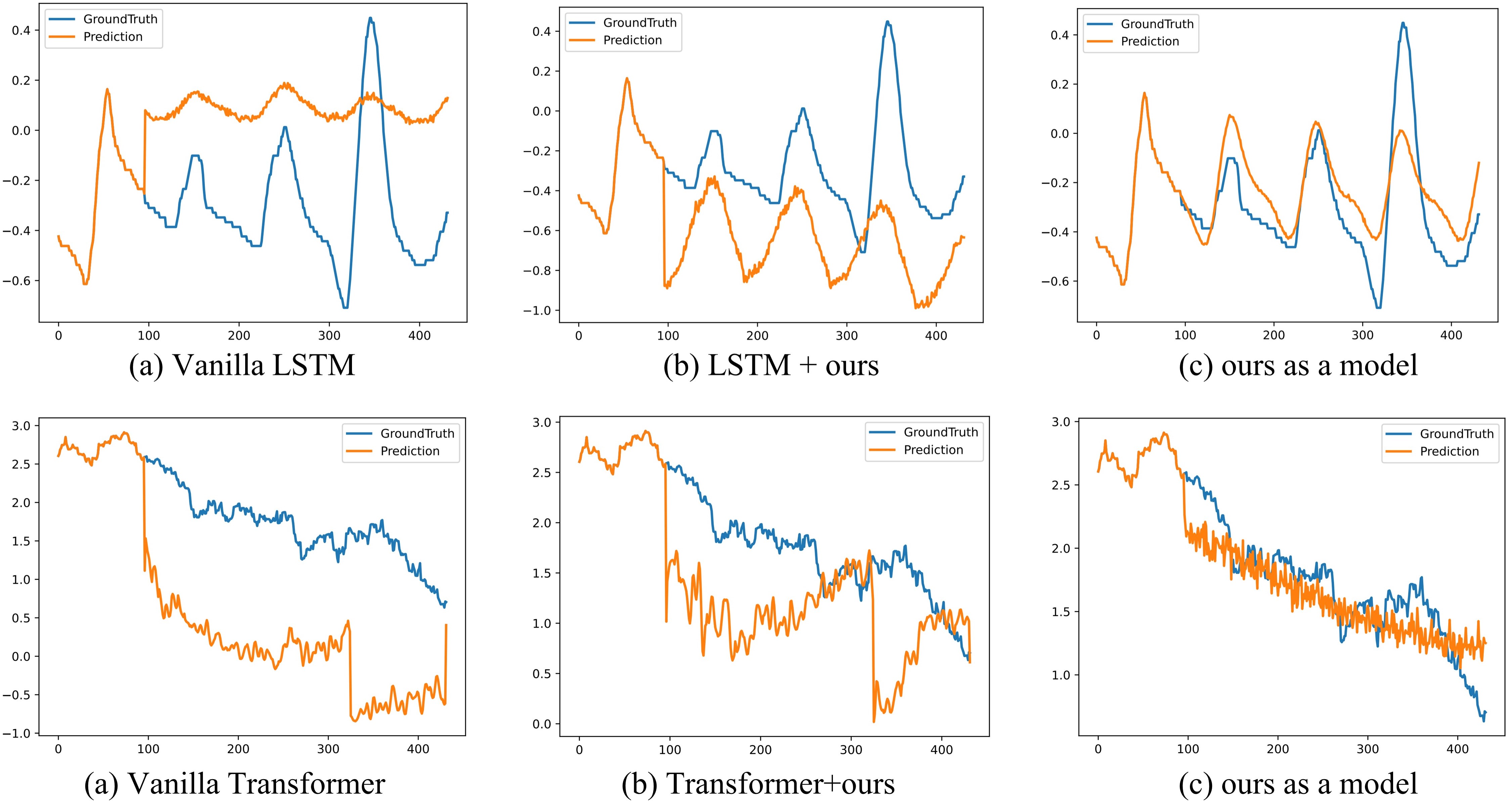}
\caption{Visualization of ETTm2 and Exchange predictions given by different models.}
\label{55}
\end{figure}

\begin{figure*}[ht]
\centering
\includegraphics[width=1.0\linewidth]{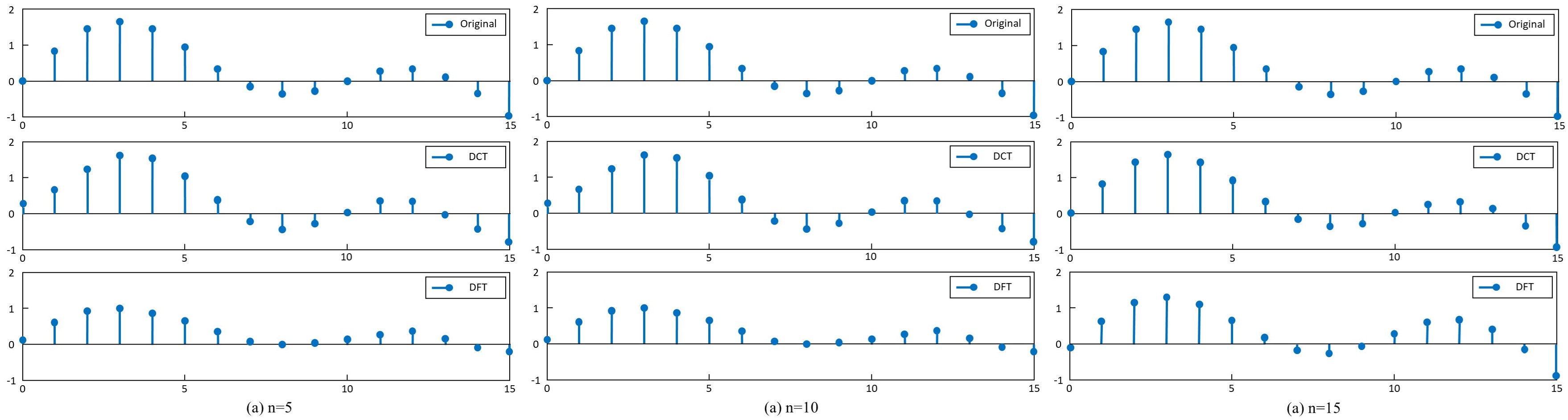}
\caption{Signal reconstruction contrast between DCT and DFT with different number of frequency components.}
\label{duibi}
\end{figure*}

\textbf{Qualitative results:} As shown in Fig.\ref{55}, we plot the prediction results of vanilla Transformer,Transformer with our FECAM block,and Our FECAM method (FECAM with a projection layer) on Exchange dataset,and plot the prediction results of vanilla LSTM, LSTM with FECAM block,and our FECAM method on ETTm2 dataset. When the input length is 96 steps and the output horizon is 336 steps, Transformer and LSTM both fail to capture the scale and bias of the future data on Exchange and ETTm2 respectively (as shown in Fig.\ref{55}(a,d)) Moreover, transformer can hardly predict a proper trend on aperiodic data such as Exchange-Rate. With Our FECAM module, both Transformer and LSTM have a better predictability compared to their vanilla version (as shown in Fig.\ref{55}(b,e)). We can see FECAM with just a projection layer can have a remarkable prediction result than other models with FECAM module, it could be the reason that parameters of FECAM is much less small than Transformer and LSTM which means Transformer and LSTM are more likely to get overfitting than FECAM with just a projection layer. These phenomena further indicate the inadequacy of existing mainstream models modelling in frequency for the TSF task. Our proposed method is beneficial for an accurate prediction of the detailed series variation, which is vital in real-world time series forecasting.

\subsection{The interpretability of FECAM}

\begin{figure}[ht]
\centering
\includegraphics[width=1.0\linewidth]{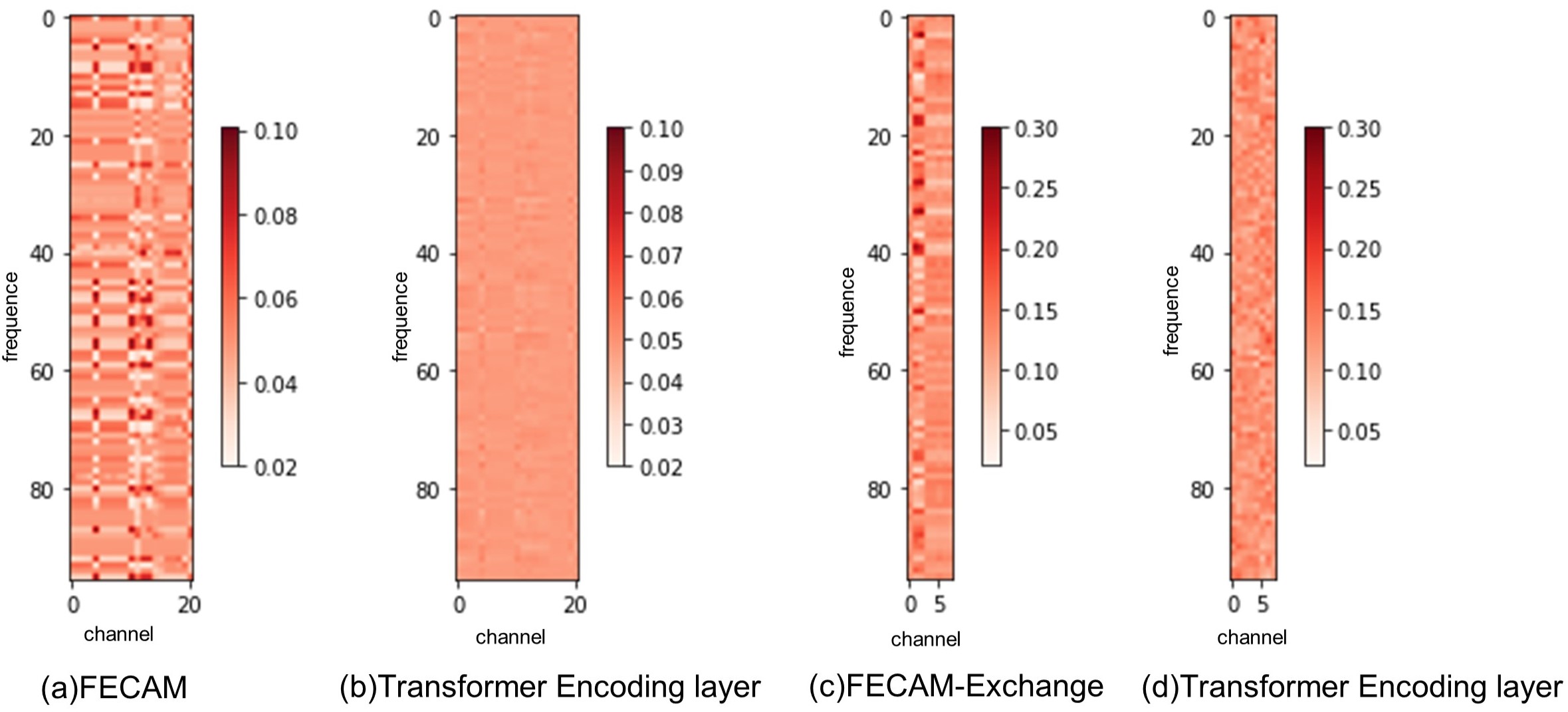}
\caption{Visualization of frequency enhanced channel attention and output tensor of encoder layer of transformer.x-axis represents channels,y-axis represents frequency from low to high,performing on datasets weather and exchange.}
\label{44}
\end{figure}

Fig.\ref{44}(a) and (b) visualize the tensor of channel attention in FECAM and encoder layer of Transformer on the weather dataset, and Fig.\ref{44}(c) and (d) visualize the tensor of channel attention in FECAM and encoder layer of Transformer on the exchange dataset. we can see FECAM can extract the importance of different channels and the significance of different frequencies with obvious patterns compared with output tensor of encoder layer in transformer.

\subsection{Energy Compaction}\label{sec4.4}
For a signal with 16 sampling points, DCT and DFT are respectively used for reconstruction. During reconstruction,DCT and DFT use $n =\{5,10,15\}$ number of components starting from low frequency. The effect is as shown in Fig.\ref{duibi}.

This experiment intuitively verified that for a signal with more energy concentrated in the low frequency, DCT can better reconstruct the signal with using less components.So Discrete Cosine Transform is more efficient in Energy compaction than Discrete Fourier Transform.

\section{Conclusion}

This paper addresses time series forecasting from the perspective of modelling in frequency domain. Unlike previous studies that most frequency extraction method are FT-based which could bring high-frequency noise to the results due to the problematic periodity, which is known as Gibbs Phenomenon. We propose Frequency enhanced channel mechanism based on Discrete Cosine Transform could intrinsically avoid G-phenomenon, and we theoretically prove the feasibility of the method. By modeling in the frequency domain, FECAM can assign channel weights to different channels, and learn the importance of different frequencies of each channel, so as to learn the frequency domain representation of time series. In the experimental stage, we visualize the frequency domain information extracted by FECAM, which verify our conjecture and proved its validity. Most importantly, for its generalization, we design this method into a module for accessibility, which can flexibly and easily use in other mainstream model like transformer-based methods and RNNs methods, etc., with just a few lines to add. Our work achieve state-of-the-art on six real-world benchmarks. This impressive generality and performance of proposed frequency enhanced channel attention mechanism can be interesting of future research for time series forecasting.

\begin{acks}
 This work was supported by the National Key R\&D Program of China (No. 2020YFB1709702) and the National Natural Science Foundation of China (No. 62073313).
\end{acks}

%\clearpage

\bibliographystyle{ACM-Reference-Format}
\bibliography{sample}


\begin{thebibliography}{47}

%%% the Web address.

\ifx \showCODEN    \undefined \def \showCODEN     #1{\unskip}     \fi
\ifx \showDOI      \undefined \def \showDOI       #1{#1}\fi
\ifx \showISBNx    \undefined \def \showISBNx     #1{\unskip}     \fi
\ifx \showISBNxiii \undefined \def \showISBNxiii  #1{\unskip}     \fi
\ifx \showISSN     \undefined \def \showISSN      #1{\unskip}     \fi
\ifx \showLCCN     \undefined \def \showLCCN      #1{\unskip}     \fi
\ifx \shownote     \undefined \def \shownote      #1{#1}          \fi
\ifx \showarticletitle \undefined \def \showarticletitle #1{#1}   \fi
\ifx \showURL      \undefined \def \showURL       {\relax}        \fi
% invisible to TeX
\providecommand\bibfield[2]{#2}
\providecommand\bibinfo[2]{#2}
\providecommand\natexlab[1]{#1}
\providecommand\showeprint[2][]{arXiv:#2}

\bibitem[\protect\citeauthoryear{Anderson and Kendall}{Anderson and
  Kendall}{1976}]%
        {Anderson1976TimeSeries2E}
\bibfield{author}{\bibinfo{person}{O. Anderson} {and} \bibinfo{person}{M.
  Kendall}.} \bibinfo{year}{1976}\natexlab{}.
\newblock \showarticletitle{Time-Series. 2nd edn.}
\newblock \bibinfo{journal}{\emph{J. R. Stat. Soc. (Series D)}}
  (\bibinfo{year}{1976}).
\newblock


\bibitem[\protect\citeauthoryear{Barra, Carta, Corriga, Podda, and
  Recupero}{Barra et~al\mbox{.}}{2020}]%
        {barra2020deep}
\bibfield{author}{\bibinfo{person}{Silvio Barra},
  \bibinfo{person}{Salvatore~Mario Carta}, \bibinfo{person}{Andrea Corriga},
  \bibinfo{person}{Alessandro~Sebastian Podda}, {and}
  \bibinfo{person}{Diego~Reforgiato Recupero}.}
  \bibinfo{year}{2020}\natexlab{}.
\newblock \showarticletitle{Deep learning and time series-to-image encoding for
  financial forecasting}.
\newblock \bibinfo{journal}{\emph{IEEE/CAA Journal of Automatica Sinica}}
  \bibinfo{volume}{7}, \bibinfo{number}{3} (\bibinfo{year}{2020}),
  \bibinfo{pages}{683--692}.
\newblock


\bibitem[\protect\citeauthoryear{Cao, Li, and Li}{Cao et~al\mbox{.}}{2019}]%
        {cao2019financial}
\bibfield{author}{\bibinfo{person}{Jian Cao}, \bibinfo{person}{Zhi Li}, {and}
  \bibinfo{person}{Jian Li}.} \bibinfo{year}{2019}\natexlab{}.
\newblock \showarticletitle{Financial time series forecasting model based on
  CEEMDAN and LSTM}.
\newblock \bibinfo{journal}{\emph{Physica A: Statistical mechanics and its
  applications}}  \bibinfo{volume}{519} (\bibinfo{year}{2019}),
  \bibinfo{pages}{127--139}.
\newblock


\bibitem[\protect\citeauthoryear{Challu, Olivares, Oreshkin, Garza,
  Mergenthaler, and Dubrawski}{Challu et~al\mbox{.}}{2022}]%
        {challu2022n}
\bibfield{author}{\bibinfo{person}{Cristian Challu}, \bibinfo{person}{Kin~G
  Olivares}, \bibinfo{person}{Boris~N Oreshkin}, \bibinfo{person}{Federico
  Garza}, \bibinfo{person}{Max Mergenthaler}, {and} \bibinfo{person}{Artur
  Dubrawski}.} \bibinfo{year}{2022}\natexlab{}.
\newblock \showarticletitle{N-HiTS: Neural Hierarchical Interpolation for Time
  Series Forecasting}.
\newblock \bibinfo{journal}{\emph{arXiv preprint arXiv:2201.12886}}
  (\bibinfo{year}{2022}).
\newblock


\bibitem[\protect\citeauthoryear{Chen, Lu, Rajeswaran, Lee, Grover, Laskin,
  Abbeel, Srinivas, and Mordatch}{Chen et~al\mbox{.}}{2021}]%
        {chen2021decisiontransformer}
\bibfield{author}{\bibinfo{person}{Lili Chen}, \bibinfo{person}{Kevin Lu},
  \bibinfo{person}{Aravind Rajeswaran}, \bibinfo{person}{Kimin Lee},
  \bibinfo{person}{Aditya Grover}, \bibinfo{person}{Michael Laskin},
  \bibinfo{person}{Pieter Abbeel}, \bibinfo{person}{Aravind Srinivas}, {and}
  \bibinfo{person}{Igor Mordatch}.} \bibinfo{year}{2021}\natexlab{}.
\newblock \showarticletitle{Decision Transformer: Reinforcement Learning via
  Sequence Modeling}.
\newblock \bibinfo{journal}{\emph{NeurIPS}} (\bibinfo{year}{2021}).
\newblock


\bibitem[\protect\citeauthoryear{Devlin, Chang, Lee, and Toutanova}{Devlin
  et~al\mbox{.}}{2019}]%
        {Devlin2019BERTPO}
\bibfield{author}{\bibinfo{person}{J. Devlin}, \bibinfo{person}{Ming-Wei
  Chang}, \bibinfo{person}{Kenton Lee}, {and} \bibinfo{person}{Kristina
  Toutanova}.} \bibinfo{year}{2019}\natexlab{}.
\newblock \showarticletitle{BERT: Pre-training of Deep Bidirectional
  Transformers for Language Understanding}. In
  \bibinfo{booktitle}{\emph{NAACL-HLT}}.
\newblock


\bibitem[\protect\citeauthoryear{Ding, Wu, Sun, Guo, and Guo}{Ding
  et~al\mbox{.}}{2020}]%
        {ding2020hierarchical}
\bibfield{author}{\bibinfo{person}{Qianggang Ding}, \bibinfo{person}{Sifan Wu},
  \bibinfo{person}{Hao Sun}, \bibinfo{person}{Jiadong Guo}, {and}
  \bibinfo{person}{Jian Guo}.} \bibinfo{year}{2020}\natexlab{}.
\newblock \showarticletitle{Hierarchical Multi-Scale Gaussian Transformer for
  Stock Movement Prediction.}. In \bibinfo{booktitle}{\emph{IJCAI}}.
  \bibinfo{pages}{4640--4646}.
\newblock


\bibitem[\protect\citeauthoryear{Dosovitskiy, Beyer, Kolesnikov, Weissenborn,
  Zhai, Unterthiner, Dehghani, Minderer, Heigold, Gelly, Uszkoreit, and
  Houlsby}{Dosovitskiy et~al\mbox{.}}{2021}]%
        {dosovitskiy2021an}
\bibfield{author}{\bibinfo{person}{Alexey Dosovitskiy}, \bibinfo{person}{Lucas
  Beyer}, \bibinfo{person}{Alexander Kolesnikov}, \bibinfo{person}{Dirk
  Weissenborn}, \bibinfo{person}{Xiaohua Zhai}, \bibinfo{person}{Thomas
  Unterthiner}, \bibinfo{person}{Mostafa Dehghani}, \bibinfo{person}{Matthias
  Minderer}, \bibinfo{person}{Georg Heigold}, \bibinfo{person}{Sylvain Gelly},
  \bibinfo{person}{Jakob Uszkoreit}, {and} \bibinfo{person}{Neil Houlsby}.}
  \bibinfo{year}{2021}\natexlab{}.
\newblock \showarticletitle{An Image is Worth 16x16 Words: Transformers for
  Image Recognition at Scale}. In \bibinfo{booktitle}{\emph{ICLR}}.
\newblock
\urldef\tempurl%
\url{https://openreview.net/forum?id=YicbFdNTTy}
\showURL{%
\tempurl}


\bibitem[\protect\citeauthoryear{Foster and Richards}{Foster and
  Richards}{1991}]%
        {foster1991gibbs}
\bibfield{author}{\bibinfo{person}{J Foster} {and} \bibinfo{person}{FB
  Richards}.} \bibinfo{year}{1991}\natexlab{}.
\newblock \showarticletitle{The Gibbs phenomenon for piecewise-linear
  approximation}.
\newblock \bibinfo{journal}{\emph{The American Mathematical Monthly}}
  \bibinfo{volume}{98}, \bibinfo{number}{1} (\bibinfo{year}{1991}),
  \bibinfo{pages}{47--49}.
\newblock


\bibitem[\protect\citeauthoryear{Ghassemi, Pimentel, Naumann, Brennan, Clifton,
  Szolovits, and Feng}{Ghassemi et~al\mbox{.}}{2015}]%
        {ghassemi2015multivariate}
\bibfield{author}{\bibinfo{person}{Marzyeh Ghassemi}, \bibinfo{person}{Marco
  Pimentel}, \bibinfo{person}{Tristan Naumann}, \bibinfo{person}{Thomas
  Brennan}, \bibinfo{person}{David Clifton}, \bibinfo{person}{Peter Szolovits},
  {and} \bibinfo{person}{Mengling Feng}.} \bibinfo{year}{2015}\natexlab{}.
\newblock \showarticletitle{A multivariate timeseries modeling approach to
  severity of illness assessment and forecasting in ICU with sparse,
  heterogeneous clinical data}. In \bibinfo{booktitle}{\emph{Proceedings of the
  AAAI conference on artificial intelligence}}, Vol.~\bibinfo{volume}{29}.
\newblock


\bibitem[\protect\citeauthoryear{Gneiting and Raftery}{Gneiting and
  Raftery}{2005}]%
        {gneiting2005weather}
\bibfield{author}{\bibinfo{person}{Tilmann Gneiting} {and}
  \bibinfo{person}{Adrian~E Raftery}.} \bibinfo{year}{2005}\natexlab{}.
\newblock \showarticletitle{Weather forecasting with ensemble methods}.
\newblock \bibinfo{journal}{\emph{Science}} \bibinfo{volume}{310},
  \bibinfo{number}{5746} (\bibinfo{year}{2005}), \bibinfo{pages}{248--249}.
\newblock


\bibitem[\protect\citeauthoryear{Greff, Srivastava, Koutn{\'\i}k, Steunebrink,
  and Schmidhuber}{Greff et~al\mbox{.}}{2016}]%
        {greff2016lstm}
\bibfield{author}{\bibinfo{person}{Klaus Greff}, \bibinfo{person}{Rupesh~K
  Srivastava}, \bibinfo{person}{Jan Koutn{\'\i}k}, \bibinfo{person}{Bas~R
  Steunebrink}, {and} \bibinfo{person}{J{\"u}rgen Schmidhuber}.}
  \bibinfo{year}{2016}\natexlab{}.
\newblock \showarticletitle{LSTM: A search space odyssey}.
\newblock \bibinfo{journal}{\emph{IEEE transactions on neural networks and
  learning systems}} \bibinfo{volume}{28}, \bibinfo{number}{10}
  (\bibinfo{year}{2016}), \bibinfo{pages}{2222--2232}.
\newblock


\bibitem[\protect\citeauthoryear{Grover, Kapoor, and Horvitz}{Grover
  et~al\mbox{.}}{2015}]%
        {grover2015deep}
\bibfield{author}{\bibinfo{person}{Aditya Grover}, \bibinfo{person}{Ashish
  Kapoor}, {and} \bibinfo{person}{Eric Horvitz}.}
  \bibinfo{year}{2015}\natexlab{}.
\newblock \showarticletitle{A deep hybrid model for weather forecasting}. In
  \bibinfo{booktitle}{\emph{Proceedings of the 21th ACM SIGKDD international
  conference on knowledge discovery and data mining}}.
  \bibinfo{pages}{379--386}.
\newblock


\bibitem[\protect\citeauthoryear{Gupta, Xiao, and Bogdan}{Gupta
  et~al\mbox{.}}{2021}]%
        {Multiwavelet-based-Operator-Learning}
\bibfield{author}{\bibinfo{person}{Gaurav Gupta}, \bibinfo{person}{Xiongye
  Xiao}, {and} \bibinfo{person}{Paul Bogdan}.} \bibinfo{year}{2021}\natexlab{}.
\newblock \bibinfo{title}{Multiwavelet-based Operator Learning for Differential
  Equations}.
\newblock
\newblock
\showeprint[arxiv]{2109.13459}~[cs.LG]


\bibitem[\protect\citeauthoryear{Hu, Shen, and Sun}{Hu et~al\mbox{.}}{2018}]%
        {hu2018senet}
\bibfield{author}{\bibinfo{person}{Jie Hu}, \bibinfo{person}{Li Shen}, {and}
  \bibinfo{person}{Gang Sun}.} \bibinfo{year}{2018}\natexlab{}.
\newblock \showarticletitle{Squeeze-and-Excitation Networks}.
\newblock \bibinfo{journal}{\emph{IEEE Conference on Computer Vision and
  Pattern Recognition}}.
\newblock


\bibitem[\protect\citeauthoryear{Kandula, Yamana, Pei, Yang, Morita, and
  Shaman}{Kandula et~al\mbox{.}}{2018}]%
        {kandula2018evaluation}
\bibfield{author}{\bibinfo{person}{Sasikiran Kandula}, \bibinfo{person}{Teresa
  Yamana}, \bibinfo{person}{Sen Pei}, \bibinfo{person}{Wan Yang},
  \bibinfo{person}{Haruka Morita}, {and} \bibinfo{person}{Jeffrey Shaman}.}
  \bibinfo{year}{2018}\natexlab{}.
\newblock \showarticletitle{Evaluation of mechanistic and statistical methods
  in forecasting influenza-like illness}.
\newblock \bibinfo{journal}{\emph{Journal of The Royal Society Interface}}
  \bibinfo{volume}{15}, \bibinfo{number}{144} (\bibinfo{year}{2018}),
  \bibinfo{pages}{20180174}.
\newblock


\bibitem[\protect\citeauthoryear{Kingma and Ba}{Kingma and Ba}{2015}]%
        {DBLP:journals/corr/KingmaB14}
\bibfield{author}{\bibinfo{person}{Diederik~P. Kingma} {and}
  \bibinfo{person}{Jimmy Ba}.} \bibinfo{year}{2015}\natexlab{}.
\newblock \showarticletitle{Adam: {A} Method for Stochastic Optimization}. In
  \bibinfo{booktitle}{\emph{ICLR}}.
\newblock
\urldef\tempurl%
\url{http://arxiv.org/abs/1412.6980}
\showURL{%
\tempurl}


\bibitem[\protect\citeauthoryear{Kitaev, Kaiser, and Levskaya}{Kitaev
  et~al\mbox{.}}{2020}]%
        {kitaev2020reformer}
\bibfield{author}{\bibinfo{person}{Nikita Kitaev}, \bibinfo{person}{Lukasz
  Kaiser}, {and} \bibinfo{person}{Anselm Levskaya}.}
  \bibinfo{year}{2020}\natexlab{}.
\newblock \showarticletitle{Reformer: The Efficient Transformer}. In
  \bibinfo{booktitle}{\emph{ICLR}}.
\newblock
\urldef\tempurl%
\url{https://openreview.net/forum?id=rkgNKkHtvB}
\showURL{%
\tempurl}


\bibitem[\protect\citeauthoryear{Lai, Chang, Yang, and Liu}{Lai
  et~al\mbox{.}}{2018}]%
        {2018Modeling}
\bibfield{author}{\bibinfo{person}{Guokun Lai}, \bibinfo{person}{Wei-Cheng
  Chang}, \bibinfo{person}{Yiming Yang}, {and} \bibinfo{person}{Hanxiao Liu}.}
  \bibinfo{year}{2018}\natexlab{}.
\newblock \showarticletitle{Modeling long-and short-term temporal patterns with
  deep neural networks}. In \bibinfo{booktitle}{\emph{SIGIR}}.
\newblock


\bibitem[\protect\citeauthoryear{Li, Jin, Xuan, Zhou, Chen, Wang, and Yan}{Li
  et~al\mbox{.}}{2019}]%
        {2019Enhancing}
\bibfield{author}{\bibinfo{person}{Shiyang Li}, \bibinfo{person}{Xiaoyong Jin},
  \bibinfo{person}{Yao Xuan}, \bibinfo{person}{Xiyou Zhou},
  \bibinfo{person}{Wenhu Chen}, \bibinfo{person}{Yu-Xiang Wang}, {and}
  \bibinfo{person}{Xifeng Yan}.} \bibinfo{year}{2019}\natexlab{}.
\newblock \showarticletitle{Enhancing the Locality and Breaking the Memory
  Bottleneck of Transformer on Time Series Forecasting}. In
  \bibinfo{booktitle}{\emph{NeurIPS}}.
\newblock
\urldef\tempurl%
\url{https://proceedings.neurips.cc/paper/2019/file/6775a0635c302542da2c32aa19d86be0-Paper.pdf}
\showURL{%
\tempurl}


\bibitem[\protect\citeauthoryear{Liu, Zeng, Chen, Xu, Lai, Ma, and Xu}{Liu
  et~al\mbox{.}}{2022}]%
        {liu2022SCINet}
\bibfield{author}{\bibinfo{person}{Minhao Liu}, \bibinfo{person}{Ailing Zeng},
  \bibinfo{person}{Muxi Chen}, \bibinfo{person}{Zhijian Xu},
  \bibinfo{person}{Qiuxia Lai}, \bibinfo{person}{Lingna Ma}, {and}
  \bibinfo{person}{Qiang Xu}.} \bibinfo{year}{2022}\natexlab{}.
\newblock \showarticletitle{SCINet: Time Series Modeling and Forecasting with
  Sample Convolution and Interaction}.
\newblock \bibinfo{journal}{\emph{Thirty-sixth Conference on Neural Information
  Processing Systems (NeurIPS), 2022}} (\bibinfo{year}{2022}).
\newblock


\bibitem[\protect\citeauthoryear{Liu, Yu, Liao, Li, Lin, Liu, and Dustdar}{Liu
  et~al\mbox{.}}{2021b}]%
        {liu2021pyraformer}
\bibfield{author}{\bibinfo{person}{Shizhan Liu}, \bibinfo{person}{Hang Yu},
  \bibinfo{person}{Cong Liao}, \bibinfo{person}{Jianguo Li},
  \bibinfo{person}{Weiyao Lin}, \bibinfo{person}{Alex~X Liu}, {and}
  \bibinfo{person}{Schahram Dustdar}.} \bibinfo{year}{2021}\natexlab{b}.
\newblock \showarticletitle{Pyraformer: Low-complexity pyramidal attention for
  long-range time series modeling and forecasting}. In
  \bibinfo{booktitle}{\emph{ICLR}}.
\newblock


\bibitem[\protect\citeauthoryear{Liu, Lin, Cao, Hu, Wei, Zhang, Lin, and
  Guo}{Liu et~al\mbox{.}}{2021a}]%
        {liu2021Swin}
\bibfield{author}{\bibinfo{person}{Ze Liu}, \bibinfo{person}{Yutong Lin},
  \bibinfo{person}{Yue Cao}, \bibinfo{person}{Han Hu}, \bibinfo{person}{Yixuan
  Wei}, \bibinfo{person}{Zheng Zhang}, \bibinfo{person}{Stephen Lin}, {and}
  \bibinfo{person}{Baining Guo}.} \bibinfo{year}{2021}\natexlab{a}.
\newblock \showarticletitle{Swin Transformer: Hierarchical Vision Transformer
  Using Shifted Windows}. In \bibinfo{booktitle}{\emph{ICCV}}.
\newblock


\bibitem[\protect\citeauthoryear{Ma, Antoniou, and Toledo}{Ma
  et~al\mbox{.}}{2020}]%
        {ma2020hybrid}
\bibfield{author}{\bibinfo{person}{Tao Ma}, \bibinfo{person}{Constantinos
  Antoniou}, {and} \bibinfo{person}{Tomer Toledo}.}
  \bibinfo{year}{2020}\natexlab{}.
\newblock \showarticletitle{Hybrid machine learning algorithm and statistical
  time series model for network-wide traffic forecast}.
\newblock \bibinfo{journal}{\emph{Transportation Research Part C: Emerging
  Technologies}}  \bibinfo{volume}{111} (\bibinfo{year}{2020}),
  \bibinfo{pages}{352--372}.
\newblock


\bibitem[\protect\citeauthoryear{Maddix, Wang, and Smola}{Maddix
  et~al\mbox{.}}{2018}]%
        {Maddix2018DeepFW}
\bibfield{author}{\bibinfo{person}{Danielle~C Maddix}, \bibinfo{person}{Yuyang
  Wang}, {and} \bibinfo{person}{Alex Smola}.} \bibinfo{year}{2018}\natexlab{}.
\newblock \showarticletitle{Deep factors with gaussian processes for
  forecasting}.
\newblock \bibinfo{journal}{\emph{arXiv preprint arXiv:1812.00098}}
  (\bibinfo{year}{2018}).
\newblock


\bibitem[\protect\citeauthoryear{Moskona, Petrushev, and Saff}{Moskona
  et~al\mbox{.}}{1995}]%
        {moskona1995gibbs}
\bibfield{author}{\bibinfo{person}{E Moskona}, \bibinfo{person}{P Petrushev},
  {and} \bibinfo{person}{EB Saff}.} \bibinfo{year}{1995}\natexlab{}.
\newblock \showarticletitle{The Gibbs phenomenon for bestL 1-trigonometric
  polynomial approximation}.
\newblock \bibinfo{journal}{\emph{Constructive Approximation}}
  \bibinfo{volume}{11}, \bibinfo{number}{3} (\bibinfo{year}{1995}),
  \bibinfo{pages}{391--416}.
\newblock


\bibitem[\protect\citeauthoryear{Oreshkin, Carpov, Chapados, and
  Bengio}{Oreshkin et~al\mbox{.}}{2019}]%
        {oreshkin2019n}
\bibfield{author}{\bibinfo{person}{Boris~N Oreshkin}, \bibinfo{person}{Dmitri
  Carpov}, \bibinfo{person}{Nicolas Chapados}, {and} \bibinfo{person}{Yoshua
  Bengio}.} \bibinfo{year}{2019}\natexlab{}.
\newblock \showarticletitle{N-{BEATS}: Neural basis expansion analysis for
  interpretable time series forecasting}.
\newblock \bibinfo{journal}{\emph{ICLR}} (\bibinfo{year}{2019}).
\newblock


\bibitem[\protect\citeauthoryear{Paszke, Gross, Massa, Lerer, Bradbury, Chanan,
  Killeen, Lin, Gimelshein, Antiga, Desmaison, K{\"o}pf, Yang, DeVito, Raison,
  Tejani, Chilamkurthy, Steiner, Fang, Bai, and Chintala}{Paszke
  et~al\mbox{.}}{2019}]%
        {Paszke2019PyTorchAI}
\bibfield{author}{\bibinfo{person}{Adam Paszke}, \bibinfo{person}{S. Gross},
  \bibinfo{person}{Francisco Massa}, \bibinfo{person}{A. Lerer},
  \bibinfo{person}{James Bradbury}, \bibinfo{person}{Gregory Chanan},
  \bibinfo{person}{Trevor Killeen}, \bibinfo{person}{Z. Lin},
  \bibinfo{person}{N. Gimelshein}, \bibinfo{person}{L. Antiga},
  \bibinfo{person}{Alban Desmaison}, \bibinfo{person}{Andreas K{\"o}pf},
  \bibinfo{person}{Edward Yang}, \bibinfo{person}{Zach DeVito},
  \bibinfo{person}{Martin Raison}, \bibinfo{person}{Alykhan Tejani},
  \bibinfo{person}{Sasank Chilamkurthy}, \bibinfo{person}{Benoit Steiner},
  \bibinfo{person}{Lu Fang}, \bibinfo{person}{Junjie Bai}, {and}
  \bibinfo{person}{Soumith Chintala}.} \bibinfo{year}{2019}\natexlab{}.
\newblock \showarticletitle{PyTorch: An Imperative Style, High-Performance Deep
  Learning Library}. In \bibinfo{booktitle}{\emph{NeurIPS}}.
\newblock


\bibitem[\protect\citeauthoryear{Rangapuram, Seeger, Gasthaus, Stella, Wang,
  and Januschowski}{Rangapuram et~al\mbox{.}}{2018}]%
        {Rangapuram2018DeepSS}
\bibfield{author}{\bibinfo{person}{Syama~Sundar Rangapuram},
  \bibinfo{person}{Matthias~W Seeger}, \bibinfo{person}{Jan Gasthaus},
  \bibinfo{person}{Lorenzo Stella}, \bibinfo{person}{Yuyang Wang}, {and}
  \bibinfo{person}{Tim Januschowski}.} \bibinfo{year}{2018}\natexlab{}.
\newblock \showarticletitle{Deep state space models for time series
  forecasting}. In \bibinfo{booktitle}{\emph{NeurIPS}}.
\newblock


\bibitem[\protect\citeauthoryear{Rasp, Dueben, Scher, Weyn, Mouatadid, and
  Thuerey}{Rasp et~al\mbox{.}}{2020}]%
        {rasp2020weatherbench}
\bibfield{author}{\bibinfo{person}{Stephan Rasp}, \bibinfo{person}{Peter~D
  Dueben}, \bibinfo{person}{Sebastian Scher}, \bibinfo{person}{Jonathan~A
  Weyn}, \bibinfo{person}{Soukayna Mouatadid}, {and} \bibinfo{person}{Nils
  Thuerey}.} \bibinfo{year}{2020}\natexlab{}.
\newblock \showarticletitle{WeatherBench: a benchmark data set for data-driven
  weather forecasting}.
\newblock \bibinfo{journal}{\emph{Journal of Advances in Modeling Earth
  Systems}} \bibinfo{volume}{12}, \bibinfo{number}{11} (\bibinfo{year}{2020}),
  \bibinfo{pages}{e2020MS002203}.
\newblock


\bibitem[\protect\citeauthoryear{Reece, Reagan, Lix, Dodds, Danforth, and
  Langer}{Reece et~al\mbox{.}}{2017}]%
        {reece2017forecasting}
\bibfield{author}{\bibinfo{person}{Andrew~G Reece}, \bibinfo{person}{Andrew~J
  Reagan}, \bibinfo{person}{Katharina~LM Lix}, \bibinfo{person}{Peter~Sheridan
  Dodds}, \bibinfo{person}{Christopher~M Danforth}, {and}
  \bibinfo{person}{Ellen~J Langer}.} \bibinfo{year}{2017}\natexlab{}.
\newblock \showarticletitle{Forecasting the onset and course of mental illness
  with Twitter data}.
\newblock \bibinfo{journal}{\emph{Scientific reports}} \bibinfo{volume}{7},
  \bibinfo{number}{1} (\bibinfo{year}{2017}), \bibinfo{pages}{1--11}.
\newblock


\bibitem[\protect\citeauthoryear{Ruiz, Rueda, Cu{\'e}llar, and Pegalajar}{Ruiz
  et~al\mbox{.}}{2018}]%
        {ruiz2018energy}
\bibfield{author}{\bibinfo{person}{Luis G~Baca Ruiz}, \bibinfo{person}{R
  Rueda}, \bibinfo{person}{Manuel~P Cu{\'e}llar}, {and} \bibinfo{person}{MC
  Pegalajar}.} \bibinfo{year}{2018}\natexlab{}.
\newblock \showarticletitle{Energy consumption forecasting based on Elman
  neural networks with evolutive optimization}.
\newblock \bibinfo{journal}{\emph{Expert Systems with Applications}}
  \bibinfo{volume}{92} (\bibinfo{year}{2018}), \bibinfo{pages}{380--389}.
\newblock


\bibitem[\protect\citeauthoryear{Salinas, Flunkert, Gasthaus, and
  Januschowski}{Salinas et~al\mbox{.}}{2020}]%
        {Flunkert2017DeepARPF}
\bibfield{author}{\bibinfo{person}{David Salinas}, \bibinfo{person}{Valentin
  Flunkert}, \bibinfo{person}{Jan Gasthaus}, {and} \bibinfo{person}{Tim
  Januschowski}.} \bibinfo{year}{2020}\natexlab{}.
\newblock \showarticletitle{Deep{AR}: Probabilistic forecasting with
  autoregressive recurrent networks}.
\newblock \bibinfo{journal}{\emph{Int. J. Forecast.}} (\bibinfo{year}{2020}).
\newblock


\bibitem[\protect\citeauthoryear{Shizgal and Jung}{Shizgal and Jung}{2003}]%
        {SHIZGAL200341}
\bibfield{author}{\bibinfo{person}{Bernie~D Shizgal} {and}
  \bibinfo{person}{Jae-Hun Jung}.} \bibinfo{year}{2003}\natexlab{}.
\newblock \showarticletitle{Towards the resolution of the Gibbs phenomena}.
\newblock \bibinfo{journal}{\emph{J. Comput. Appl. Math.}}
  \bibinfo{volume}{161}, \bibinfo{number}{1} (\bibinfo{year}{2003}),
  \bibinfo{pages}{41--65}.
\newblock
\showISSN{0377-0427}
\urldef\tempurl%
\url{https://doi.org/10.1016/S0377-0427(03)00500-4}
\showDOI{\tempurl}


\bibitem[\protect\citeauthoryear{Vaswani, Shazeer, Parmar, Uszkoreit, Jones,
  Gomez, Kaiser, and Polosukhin}{Vaswani et~al\mbox{.}}{2017}]%
        {NIPS2017_3f5ee243}
\bibfield{author}{\bibinfo{person}{Ashish Vaswani}, \bibinfo{person}{Noam
  Shazeer}, \bibinfo{person}{Niki Parmar}, \bibinfo{person}{Jakob Uszkoreit},
  \bibinfo{person}{Llion Jones}, \bibinfo{person}{Aidan~N Gomez},
  \bibinfo{person}{Lukasz Kaiser}, {and} \bibinfo{person}{Illia Polosukhin}.}
  \bibinfo{year}{2017}\natexlab{}.
\newblock \showarticletitle{Attention is All you Need}. In
  \bibinfo{booktitle}{\emph{NeurIPS}}.
\newblock
\urldef\tempurl%
\url{https://proceedings.neurips.cc/paper/2017/file/3f5ee243547dee91fbd053c1c4a845aa-Paper.pdf}
\showURL{%
\tempurl}


\bibitem[\protect\citeauthoryear{Wang, Lei, Zhang, Zhou, and Peng}{Wang
  et~al\mbox{.}}{2019}]%
        {wang2019review}
\bibfield{author}{\bibinfo{person}{Huaizhi Wang}, \bibinfo{person}{Zhenxing
  Lei}, \bibinfo{person}{Xian Zhang}, \bibinfo{person}{Bin Zhou}, {and}
  \bibinfo{person}{Jianchun Peng}.} \bibinfo{year}{2019}\natexlab{}.
\newblock \showarticletitle{A review of deep learning for renewable energy
  forecasting}.
\newblock \bibinfo{journal}{\emph{Energy Conversion and Management}}
  \bibinfo{volume}{198} (\bibinfo{year}{2019}), \bibinfo{pages}{111799}.
\newblock


\bibitem[\protect\citeauthoryear{Wei, Li, Peng, Zeng, and Lu}{Wei
  et~al\mbox{.}}{2019}]%
        {wei2019conventional}
\bibfield{author}{\bibinfo{person}{Nan Wei}, \bibinfo{person}{Changjun Li},
  \bibinfo{person}{Xiaolong Peng}, \bibinfo{person}{Fanhua Zeng}, {and}
  \bibinfo{person}{Xinqian Lu}.} \bibinfo{year}{2019}\natexlab{}.
\newblock \showarticletitle{Conventional models and artificial
  intelligence-based models for energy consumption forecasting: A review}.
\newblock \bibinfo{journal}{\emph{Journal of Petroleum Science and
  Engineering}}  \bibinfo{volume}{181} (\bibinfo{year}{2019}),
  \bibinfo{pages}{106187}.
\newblock


\bibitem[\protect\citeauthoryear{Wen, Torkkola, Narayanaswamy, and Madeka}{Wen
  et~al\mbox{.}}{2017}]%
        {Wen2017AMQ}
\bibfield{author}{\bibinfo{person}{Ruofeng Wen}, \bibinfo{person}{Kari
  Torkkola}, \bibinfo{person}{Balakrishnan Narayanaswamy}, {and}
  \bibinfo{person}{Dhruv Madeka}.} \bibinfo{year}{2017}\natexlab{}.
\newblock \showarticletitle{A multi-horizon quantile recurrent forecaster}.
\newblock \bibinfo{journal}{\emph{NeurIPS}} (\bibinfo{year}{2017}).
\newblock


\bibitem[\protect\citeauthoryear{Wong, Walters, Jiang, Molnar, and Yu}{Wong
  et~al\mbox{.}}{2020}]%
        {wong2020traffic}
\bibfield{author}{\bibinfo{person}{Steven Wong}, \bibinfo{person}{Robin
  Walters}, \bibinfo{person}{Lejun Jiang}, \bibinfo{person}{Tamas~G Molnar},
  {and} \bibinfo{person}{Rose Yu}.} \bibinfo{year}{2020}\natexlab{}.
\newblock \showarticletitle{Traffic Forecasting using Vehicle-to-Vehicle
  Communication and Recurrent Neural Networks}.
\newblock \bibinfo{journal}{\emph{NIPS., Dec}} (\bibinfo{year}{2020}).
\newblock


\bibitem[\protect\citeauthoryear{Wu, Xu, Wang, and Long}{Wu
  et~al\mbox{.}}{2021b}]%
        {wu2021autoformer}
\bibfield{author}{\bibinfo{person}{Haixu Wu}, \bibinfo{person}{Jiehui Xu},
  \bibinfo{person}{Jianmin Wang}, {and} \bibinfo{person}{Mingsheng Long}.}
  \bibinfo{year}{2021}\natexlab{b}.
\newblock \showarticletitle{Autoformer: Decomposition Transformers with
  {Auto-Correlation} for Long-Term Series Forecasting}. In
  \bibinfo{booktitle}{\emph{NeurIPS}}.
\newblock


\bibitem[\protect\citeauthoryear{Wu, Ni, Cheng, Zong, Song, Chen, Liu, Zhang,
  Chen, and Davidson}{Wu et~al\mbox{.}}{2021a}]%
        {wu2021dynamic}
\bibfield{author}{\bibinfo{person}{Yinjun Wu}, \bibinfo{person}{Jingchao Ni},
  \bibinfo{person}{Wei Cheng}, \bibinfo{person}{Bo Zong},
  \bibinfo{person}{Dongjin Song}, \bibinfo{person}{Zhengzhang Chen},
  \bibinfo{person}{Yanchi Liu}, \bibinfo{person}{Xuchao Zhang},
  \bibinfo{person}{Haifeng Chen}, {and} \bibinfo{person}{Susan~B Davidson}.}
  \bibinfo{year}{2021}\natexlab{a}.
\newblock \showarticletitle{Dynamic Gaussian mixture based deep generative
  model for robust forecasting on sparse multivariate time series}. In
  \bibinfo{booktitle}{\emph{Proceedings of the AAAI Conference on Artificial
  Intelligence}}, Vol.~\bibinfo{volume}{35}. \bibinfo{pages}{651--659}.
\newblock


\bibitem[\protect\citeauthoryear{Xu, Qin, Sun, Wang, Chen, and Ren}{Xu
  et~al\mbox{.}}{2020}]%
        {Xu_2020_CVPR}
\bibfield{author}{\bibinfo{person}{Kai Xu}, \bibinfo{person}{Minghai Qin},
  \bibinfo{person}{Fei Sun}, \bibinfo{person}{Yuhao Wang},
  \bibinfo{person}{Yen-Kuang Chen}, {and} \bibinfo{person}{Fengbo Ren}.}
  \bibinfo{year}{2020}\natexlab{}.
\newblock \showarticletitle{Learning in the Frequency Domain}. In
  \bibinfo{booktitle}{\emph{Proceedings of the IEEE/CVF Conference on Computer
  Vision and Pattern Recognition (CVPR)}}.
\newblock


\bibitem[\protect\citeauthoryear{Yu, Zheng, Anandkumar, and Yue}{Yu
  et~al\mbox{.}}{2017}]%
        {2017Long}
\bibfield{author}{\bibinfo{person}{Rose Yu}, \bibinfo{person}{Stephan Zheng},
  \bibinfo{person}{Anima Anandkumar}, {and} \bibinfo{person}{Yisong Yue}.}
  \bibinfo{year}{2017}\natexlab{}.
\newblock \showarticletitle{Long-term forecasting using tensor-train RNNs}.
\newblock \bibinfo{journal}{\emph{arXiv preprint arXiv:1711.00073}}
  (\bibinfo{year}{2017}).
\newblock


\bibitem[\protect\citeauthoryear{Zeng, Hu, Zhou, Li, Liu, and Liu}{Zeng
  et~al\mbox{.}}{2022}]%
        {zeng2022muformer}
\bibfield{author}{\bibinfo{person}{Pengyu Zeng}, \bibinfo{person}{Guoliang Hu},
  \bibinfo{person}{Xiaofeng Zhou}, \bibinfo{person}{Shuai Li},
  \bibinfo{person}{Pengjie Liu}, {and} \bibinfo{person}{Shurui Liu}.}
  \bibinfo{year}{2022}\natexlab{}.
\newblock \showarticletitle{Muformer: A long sequence time-series forecasting
  model based on modified multi-head attention}.
\newblock \bibinfo{journal}{\emph{Knowledge-Based Systems}}
  \bibinfo{volume}{254} (\bibinfo{year}{2022}), \bibinfo{pages}{109584}.
\newblock


\bibitem[\protect\citeauthoryear{Zhang and Patras}{Zhang and Patras}{2018}]%
        {zhang2018long}
\bibfield{author}{\bibinfo{person}{Chaoyun Zhang} {and} \bibinfo{person}{Paul
  Patras}.} \bibinfo{year}{2018}\natexlab{}.
\newblock \showarticletitle{Long-term mobile traffic forecasting using deep
  spatio-temporal neural networks}. In \bibinfo{booktitle}{\emph{Proceedings of
  the Eighteenth ACM International Symposium on Mobile Ad Hoc Networking and
  Computing}}. \bibinfo{pages}{231--240}.
\newblock


\bibitem[\protect\citeauthoryear{Zhang, Chang, Meng, Xiang, and Pan}{Zhang
  et~al\mbox{.}}{2020}]%
        {zhang2020spatio}
\bibfield{author}{\bibinfo{person}{Qi Zhang}, \bibinfo{person}{Jianlong Chang},
  \bibinfo{person}{Gaofeng Meng}, \bibinfo{person}{Shiming Xiang}, {and}
  \bibinfo{person}{Chunhong Pan}.} \bibinfo{year}{2020}\natexlab{}.
\newblock \showarticletitle{Spatio-temporal graph structure learning for
  traffic forecasting}. In \bibinfo{booktitle}{\emph{Proceedings of the AAAI
  Conference on Artificial Intelligence}}, Vol.~\bibinfo{volume}{34}.
  \bibinfo{pages}{1177--1185}.
\newblock


\bibitem[\protect\citeauthoryear{Zhou, Zhang, Peng, Zhang, Li, Xiong, and
  Zhang}{Zhou et~al\mbox{.}}{2021}]%
        {haoyietal-informer-2021}
\bibfield{author}{\bibinfo{person}{Haoyi Zhou}, \bibinfo{person}{Shanghang
  Zhang}, \bibinfo{person}{Jieqi Peng}, \bibinfo{person}{Shuai Zhang},
  \bibinfo{person}{Jianxin Li}, \bibinfo{person}{Hui Xiong}, {and}
  \bibinfo{person}{Wancai Zhang}.} \bibinfo{year}{2021}\natexlab{}.
\newblock \showarticletitle{Informer: Beyond Efficient Transformer for Long
  Sequence Time-Series Forecasting}. In \bibinfo{booktitle}{\emph{AAAI}}.
\newblock


\end{thebibliography}

\end{document}